%% file: main.tex
\documentclass[lettersize,journal]{IEEEtran}
\usepackage{amsmath,amsfonts}
\usepackage{algorithmic}
\usepackage{algorithm}
\usepackage{array}
\usepackage[caption=false,font=normalsize,labelfont=sf,textfont=sf]{subfig}
\usepackage{textcomp}
\usepackage{stfloats}
\usepackage{url}
\usepackage{verbatim}
\usepackage{graphicx}
\usepackage{cite}
\usepackage{booktabs}
\usepackage{pifont}
\usepackage{multirow}
\usepackage{makecell}
\usepackage{colortbl}
\usepackage[backref]{hyperref} 
\hypersetup{hidelinks}

\usepackage{xcolor}
\definecolor{demored}{RGB}{192, 0, 0}
\definecolor{demogreen}{RGB}{84, 130, 53}
\definecolor{demoblue}{RGB}{47, 85, 151}
\definecolor{myred}{RGB}{255, 230, 230}
\definecolor{myblue}{RGB}{229, 241, 251}

\begin{document}

\title{Multi-Dimensional Quality Assessment for AI-Generated Human-Centric Videos: Dataset and Model}

\author{Sijing Wu, Yunhao Li, Huiyu Duan, Yucheng Zhu, Xiongkuo Min, \\ Patrick Le Callet,~\IEEEmembership{Fellow,~IEEE}, and Guangtao Zhai,~\IEEEmembership{Fellow,~IEEE}
\thanks{Sijing Wu, Yunhao Li, Huiyu Duan, Xiongkuo Min, and Guangtao Zhai are with the Institute of Image Communication and Network Engineering, Shanghai Jiao Tong University, Shanghai, China (e-mail: \{wusijing, lyhsjtu, huiyuduan, minxiongkuo, zhaiguangtao\}@sjtu.edu.cn).}
\thanks{Yucheng Zhu is with the USC-SJTU Institute of Cultural and Creative Industry, Shanghai Jiao Tong University, Shanghai, China (e-mail: zyc420@sjtu.edu.cn).}
\thanks{Patrick Le Callet is with the Polytech Nantes, Universit\'{e} de Nantes, France (e-mail: patrick.lecallet@univ-nantes.fr).}
}



\maketitle

\begin{abstract}
AI-generated human-centric videos play a crucial role in a wide range of modern applications. However, they often suffer from quality issues and semantic mismatches, underscoring the importance of effective quality assessment for such videos.
To this end, we extend our previous dataset HVEval with pairwise preference annotations, resulting in \textbf{HVEval+}, the largest holistic quality assessment dataset for AI-generated human-centric videos, which comprises 1k prompts based on a comprehensive taxonomy, 20k videos generated by 24 text-to-video (T2V) models, and extensive human annotations, including 60k mean opinion scores (MOSs) and 60k preference pairs across 3 dimensions (\textit{i.e.}, spatial quality, temporal quality, and text-video correspondence), as well as 20k category-specific question-answer (Q\&A) pairs.
Along with the HVEval+ dataset, we further propose \textbf{MoE-Rater}, a Mixture-of-Experts (MoE)-inspired and multimodal large language model (MLLM)-based all-in-one method that supports multi-dimensional quality rating, multi-dimensional pairwise comparison, and category-specific question answering within a single model. Specifically, we introduce Mixture of Projector Experts (MoPE) and Mixture of LoRA Experts (MoLE), together with a three-stage training strategy consisting of task-aware pre-training, task-specific adaptation, and adaptive routing optimization, to effectively unify multiple tasks, resulting in superior performance on both HVEval+ and Human-AGVQA datasets.
Extensive experiments and comprehensive analysis demonstrate the significant potential of the HVEval+ dataset and the MoE-Rater method in advancing AI-generated video quality assessment and further facilitating the evaluation and optimization of T2V models.

\end{abstract}


\section{Introduction}

\begin{figure}[h]
\centering
\includegraphics[width=0.98\linewidth]{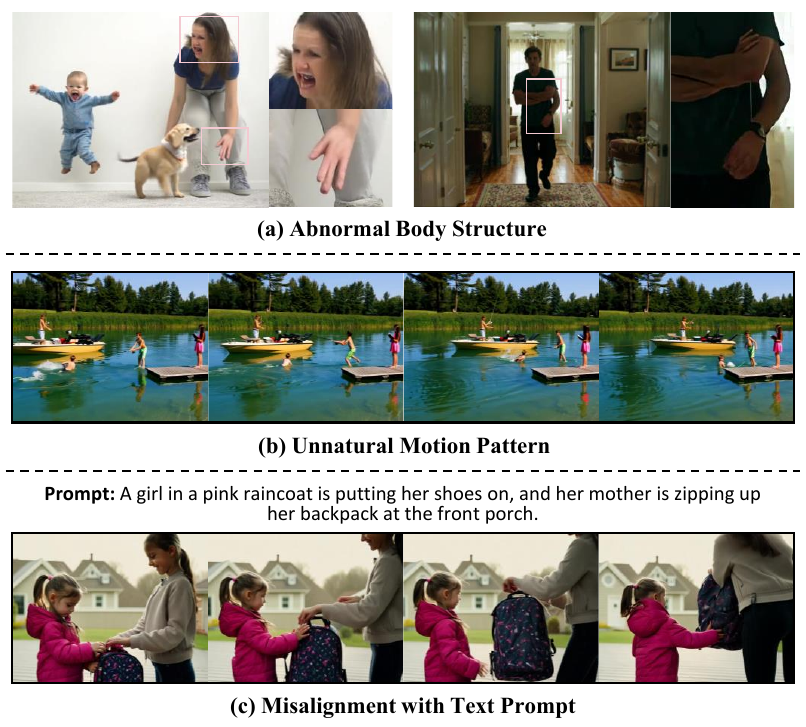}
\vspace{-1.7mm}
\caption{Illustration of typical distortions in AI-generated human-centric videos. (a) Abnormal body structures, such as melting faces, missing fingers, and extra arms. (b) Unnatural motion patterns, including anatomically and physically implausible motions. (c) Misalignment with the text prompt, for example, inconsistencies in human attributes or actions.}
\vspace{-2.3mm}
\label{fig:distortion}
\end{figure}

\IEEEPARstart{R}{ecent} advances in text-to-video (T2V) generation \cite{kong2024hunyuanvideo,wan2025wan,teng2025magi,sora,PixVerse,pika} have made noticeable progress in creating diverse and realistic videos from natural-language prompts, offering unprecedented opportunities for numerous applications such as automated content creation, education, and digital entertainment.
Among them, human-centric videos are particularly important, as they constitute the core of most real-world applications, exhibit greater visual and temporal complexity, and exert a stronger influence on human perception \cite{li2025openhumanvid,zhao2025humanomni,wu2025singinghead,zhang2025human,wu2025hveval}.
However, current T2V models often struggle to generate high-quality human-centric videos, frequently exhibiting issues such as abnormal body structures, unnatural motion patterns, and misalignment with the given text prompts \cite{wang2024generated,zhang2025human,wu2025hveval}, as illustrated in Figure \ref{fig:distortion}.
Therefore, it is important to thoroughly investigate how to effectively evaluate the quality of these videos, as such evaluation is essential for monitoring video quality in T2V applications, assessing and comparing T2V models, and providing feedback to further optimize them.

Current evaluation methods primarily assess the generation quality of general-purpose videos and can be broadly categorized into objective metrics, benchmark-based evaluation, and human-aligned evaluation.
Objective metrics such as Inception Score (IS) \cite{salimans2016improved}, Fréchet Inception Distance (FID) \cite{heusel2017gans}, and Fréchet Video Distance (FVD) \cite{unterthiner2018towards} typically evaluate generated videos through statistical features and distribution distances, which often fail to align with human perceptual judgments.
Benchmark-based evaluation frameworks, exemplified by VBench series \cite{huang2024vbench,zheng2025vbench} and T2V-CompBench \cite{sun2024t2v}, evaluate T2V models using prompts derived from comprehensive taxonomies and intuitive evaluation metrics that tend to align with human preferences. However, these methods evaluate T2V models rather than the generated videos themselves, limiting their ability to provide effective feedback for video quality monitoring and model improvement.
Human-aligned evaluation methods \cite{kou2024subjective,zhang2024benchmarking,wang2025aigv} generally learn a neural network to regress video quality scores from human-annotated datasets, which makes them inherently aligned with human perception and suitable for serving as feedback.
However, most existing methods focus on general-purpose videos, largely due to the lack of large-scale and comprehensive quality assessment datasets for AI-generated human-centric videos.

\input{tabs/dataset}

To fill this gap, we present \textbf{HVEval+}, a holistic quality assessment dataset for AI-generated human-centric videos, developed as an extension of our previous dataset, HVEval \cite{wu2025hveval} (See Table \ref{tab:dataset}).
HVEval+ consists of (1) 1k prompts based on a comprehensive taxonomy, (2) 20k videos generated by 24 popular T2V models, and (3) extensive human annotations, including 60k MOSs and 60k preference pairs across 3 dimensions (\textit{i.e.}, spatial quality, temporal quality, and text-video correspondence), as well as 20k category-specific Q\&A pairs.
Specifically, we first design a comprehensive taxonomy that organizes human-related prompts into 7 major categories and 22 subcategories. Based on the taxonomy, we incorporate internal thinking \cite{peng2024dreambench++} and human-in-the-loop strategies to construct a prompt suite covering all categories using ChatGPT-4o \cite{chatgpt4o}.
We then generate 20,000 videos from the 1,000 prompts using 24 T2V models.
To mitigate the inherent noise and ambiguity of absolute MOS annotations and to better capture relative perceptual judgments that more faithfully reflect human preferences and subtle quality differences, we further construct 60,000 video pairs, each formed from samples generated under the same prompt.
Finally, we recruit 33 subjects to annotate the data with quality scores and pairwise preferences across three dimensions, along with category-specific yes/no questions.

Along with the HVEval+ dataset, we propose an MoE-inspired all-in-one method, termed \textbf{MoE-Rater}, that enables multi-dimensional quality rating, multi-dimensional pairwise comparison, and category-specific question answering in a single model.
Specifically, MoE-Rater enhances multimodal large language models (MLLMs) with temporal features to capture comprehensive temporal information, and further incorporates Mixture of Projector Experts (MoPE) and Mixture of LoRA Experts (MoLE) modules together with a three-stage training strategy to unify multiple tasks.
The spatial and temporal features are extracted by vision and temporal encoders, projected through separate MoPE modules, namely S-MoPE and T-MoPE, and then fed into a pre-trained large language model (LLM) equipped with MoLE, together with task-specific textual instructions, for multimodal feature fusion and reasoning to support multi-task responses.
The S-MoPE and T-MoPE modules each consist of multiple task-specific MLP projectors with independent gating networks for top-k expert activation and weighted aggregation, while the MoLE module consists of multiple task-specific LoRA experts with its own gating network.
MoE-Rater is trained in three stages, including task-aware pre-training across all tasks, task-specific adaptation of experts for each task, and adaptive routing optimization to aggregate multiple experts.
Extensive experiments demonstrate the effectiveness of MoE-Rater across all tasks and its superiority over state-of-the-art methods.

In summary, the main contributions of this paper are:
\begin{itemize}

    \item We extend HVEval to HVEval+ by introducing pairwise preference annotations, resulting in the largest holistic quality assessment dataset for AI-generated human-centric videos, with 1k prompts, 20k videos, 60k quality scores and 60k preference pairs across three dimensions, and 20k category-specific Q\&A pairs.

    \item Based on the HVEval+ dataset, we benchmark VQA metrics, vision–language alignment metrics, and MLLMs across multiple evaluation tasks.

    \item We propose \textbf{MoE-Rater}, an MoE-inspired and MLLM-based all-in-one method for multi-dimensional quality rating, multi-dimensional pairwise comparison, and category-specific question answering.

    \item The proposed S-MoPE, T-MoPE, and MoLE modules, together with a three-stage training strategy, allow MoE-Rater to support all tasks within a single unified model.

    \item Extensive experiments on both HVEval+ and Human-AGVQA datasets demonstrate the superiority of the proposed MoE-Rater method.
\end{itemize}

\section{Related Work}
\subsection{Text-to-video Generation}
Text-to-video (T2V) generation aims to synthesize temporally coherent video sequences that are semantically aligned with given textual descriptions, and has received increasing attention in recent years due to its numerous applications.
With the rapid progress of diffusion models, diffusion-based T2V models have become a mainstream paradigm, demonstrating substantial improvements in visual quality and training stability compared to earlier video generation methods.
Popular T2V models can be broadly divided into three categories, including diffusion U-Net-based models \cite{lin2024animatediff,wang2024animatelcm,yuan2025magictime,wang2023modelscope,zhang2024show,li2024t2v,chen2024videocrafter2,Zeroscope}, diffusion transformer (DiT)-based models \cite{yang2024cogvideox,kong2024hunyuanvideo,hacohen2024ltx,ma2024latte,ma2025step,wan2025wan,genmo2024mochi}, and autoregressive models \cite{teng2025magi,chen2025skyreels}.
Diffusion U-Net-based models \cite{lin2024animatediff,wang2024animatelcm,yuan2025magictime,wang2023modelscope,zhang2024show,li2024t2v,chen2024videocrafter2,Zeroscope} typically employ a U-Net backbone to model the spatiotemporal denoising process, leveraging convolutional inductive biases to generate temporally consistent videos from textual conditions.
DiT-based models \cite{yang2024cogvideox,kong2024hunyuanvideo,hacohen2024ltx,ma2024latte,ma2025step,wan2025wan,genmo2024mochi} adopt transformer backbones in place of convolutional architectures, enabling more scalable modeling of long-range spatial and temporal dependencies via attention mechanisms.
Autoregressive models \cite{teng2025magi,chen2025skyreels} generate videos by sequentially predicting tokens or frames conditioned on previous outputs and textual inputs, explicitly modeling videos as a causal sequence of conditional distributions over time.
Despite recent advances in T2V models, they still exhibit notable limitations in human-centric video generation, underscoring the importance of reliable evaluation for generated videos and T2V models.

\subsection{Evaluation of Text-to-video Generation}
The rapid emergence of T2V models highlights the critical need for systematic evaluation of both the models themselves and the videos they generate.
Current evaluation methods can be generally divided into objective metrics \cite{salimans2016improved,heusel2017gans,unterthiner2018towards}, benchmark-based evaluation \cite{huang2024vbench,zheng2025vbench,sun2024t2v}, and human-aligned evaluation \cite{kou2024subjective,zhang2024benchmarking,wang2025aigv}.
Objective metrics \cite{salimans2016improved,heusel2017gans,unterthiner2018towards} typically evaluate generated videos via statistical feature representations and distributional distances. However, these metrics measure statistical similarity rather than perceptual quality, which often results in discrepancies with human perceptual judgments.
Benchmark-based evaluation frameworks \cite{huang2024vbench,zheng2025vbench,sun2024t2v} evaluate T2V models using prompts constructed from comprehensive taxonomies, together with a set of intuitive evaluation metrics that are often designed using visual analysis techniques \cite{zhou2025cofnet,zhou2025afes} and tend to align with human preferences. These benchmarks aim to evaluate T2V models rather than individual generated videos, limiting their ability to provide video-level perceptual feedback for monitoring generated video quality and improving T2V models.
Human-aligned evaluation methods \cite{kou2024subjective,zhang2024benchmarking,wang2025aigv} typically learn a network to regress video quality scores from human-annotated datasets and are therefore inherently aligned with human perceptual judgments, making them well suited to serve as effective feedback. However, most focus on general-purpose videos, which differ from human-centric videos in both distortion characteristics and their impact on quality of experience (QoE), as the human visual system is particularly sensitive to distortions in human regions.
Recently, Human-AGVQA \cite{zhang2025human} has shifted attention toward human-centric videos, but it mainly considers human appearance and activities while neglecting text–video correspondence. Moreover, it is limited in scale and provides only MOS annotations, restricting its applicability and generalization ability.
To this end, we present the largest and first holistic quality assessment dataset specifically designed for AI-generated human-centric videos, featuring both large-scale data and rich annotations.

\subsection{Video Quality Assessment}
Video quality assessment (VQA) is a long-standing research problem that aims to automatically predict perceptual quality scores of videos in accordance with human perception.
Building upon image quality assessment (IQA) \cite{zhou2022end,zhou2022attentional,lan2023multilevel,li2024blind,shen2024graph,li2024bridging,zhou2025blind,shen2025image,lan2025no,li2026towards,song2025uni,shi2024beyond}, VQA \cite{sun2022deep,sun2024analysis,wu2022fast,chen2022dynamic,wu2023exploring,wu2023q,lu2024kvq,liu2024scaling,yan2024video} further extends perceptual quality modeling from static images to temporal video content.
To facilitate the development of VQA methods, numerous datasets have been constructed over the past decades, which can be broadly categorized into distortion-based datasets with synthetic degradations \cite{lin2020pea265,lu2024kvq}, user-generated content (UGC) datasets capturing in-the-wild distortions \cite{ying2021patch,wu2025fvq,jin2025rgc}, and emerging datasets designed for AI-generated content \cite{kou2024subjective,zhang2024benchmarking,wang2025aigv,zhang2025human,wu2025hveval,gao2025ges,gao2026sfqa}.
Based on VQA datasets, VQA methods have been widely explored and can be roughly divided into two categories: traditional methods \cite{korhonen2019two,tu2021rapique,tu2021ugc} and learning-based methods \cite{chen2021learning,sun2022deep,shen2022end,sun2024analysis,wu2022fast,chen2022dynamic,wu2023exploring,wu2023q,lu2024kvq,liu2024scaling,yan2024video,hu2025blind,zhou2026mi3s,xu2026objective}.
Traditional methods typically rely on hand-crafted features and estimate quality scores using regression algorithms such as support vector regression (SVR). Owing to their limited representational capacity, these methods often struggle to handle the diverse and complex distortions encountered in real-world scenarios.
Over the past few years, learning-based methods have achieved superior performance by adopting data-driven paradigms and advanced network architectures.
However, these methods are limited to quality score prediction, with most producing only a single score per video, which restricts their application scenarios.
In contrast, we propose the first all-in-one model that jointly supports single-video quality score prediction and double-video pairwise comparison across multiple evaluation dimensions.

\begin{figure}[t]
\centering
\includegraphics[width=0.95\linewidth]{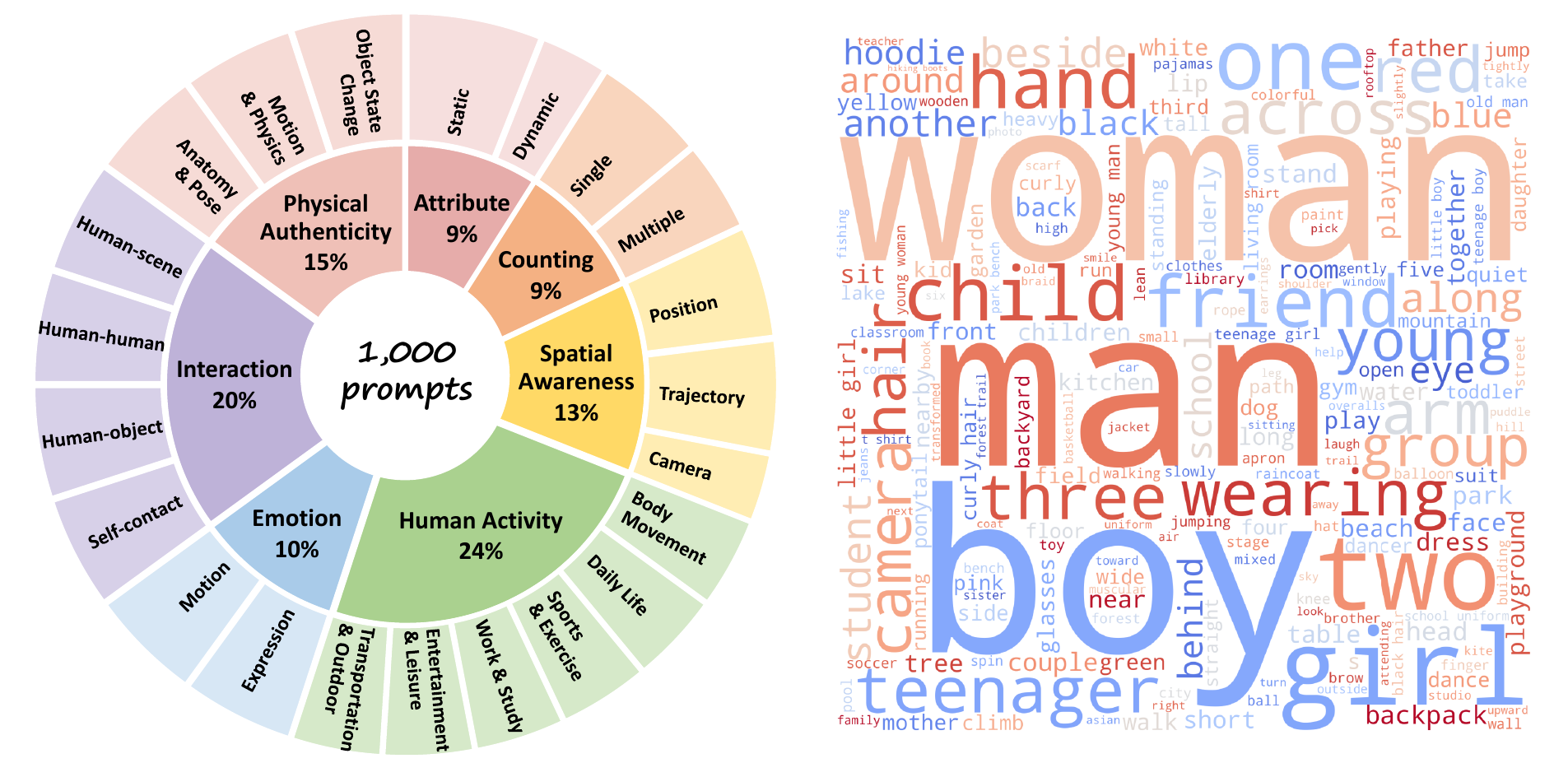}
\vspace{-1mm}
\caption{Overview of the prompt suite statistics. Left: taxonomy of the human-centric prompts in the HVEval+ dataset. Right: word cloud of the prompts.}
\vspace{-2mm}
\label{fig:prompt}
\end{figure}

\section{HVEval+ Dataset}
\begin{figure}[t]
\centering
\includegraphics[width=0.96\linewidth]{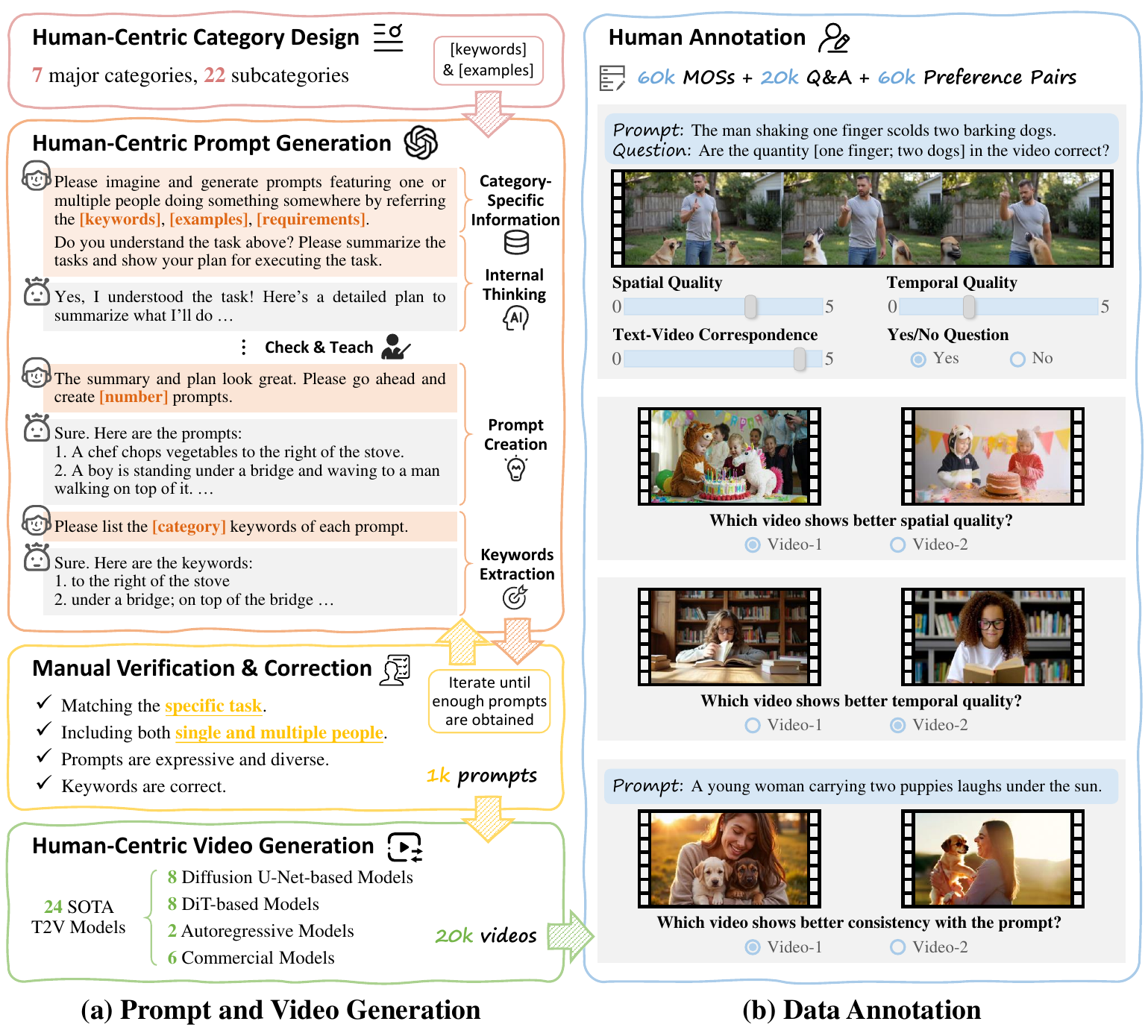}
\vspace{-1.6mm}
\caption{Dataset construction pipeline. (a) We first generate 1k prompts based on the taxonomy using ChatGPT-4o with internal thinking and human-in-the-loop strategies. Then, we use 24 T2V models to generate 20k videos, and further construct 60k video pairs from these videos. (b) We recruit 33 subjects to annotate the data with quality scores and pairwise preferences across three dimensions (\textit{i.e.}, spatial quality, temporal quality, and text-video correspondence), as well as category-specific yes/no questions.}
\vspace{-3.5mm}
\label{fig:dataset}
\end{figure}

\subsection{Taxonomy and Prompt Suite}
We generate 1,000 high-quality, human-centric prompts across 7 major categories and 22 subcategories for T2V model generation. These prompts are constructed based on a comprehensive taxonomy, leveraging ChatGPT-4o \cite{chatgpt4o} with internal thinking \cite{peng2024dreambench++} and human-in-the-loop strategies, as illustrated in Figure \ref{fig:dataset} (a).

The taxonomy is shown in Figure \ref{fig:prompt}, which organizes the human-centric prompts into 7 major categories:
\noindent{(1) \textbf{Attribute}}, which focuses on both the static and dynamic attributes of humans. Static attributes are constructed from random combinations of age, gender, race, body type, haircut, clothing, and accessory. Each attribute contains a diverse collection of commonly used keywords for selection. Dynamic attributes focus on imaginative attribute change, such as \textit{``The sickly child transformed into a vibrant superhero"}.
\noindent{(2) \textbf{Counting}}, which involves quantity descriptions for both single and multiple people. Single person counting refers to counting the attributes or actions of an individual, such as \textit{``two ponytails"}, \textit{``jumping on one foot"}, and \textit{``clapping hands three times"}. Multiple people counting focuses on enumerating the number of people present.
\noindent{(3) \textbf{Spatial awareness}}, which aims to evaluate relative positions, human motion trajectories, and camera trajectories in video generation. Relative positions are described using spatial prepositions such as \textit{``above"}, \textit{``between"}, \textit{``in front of"}, and \textit{``to the left of"}. Human motion trajectories are randomly sampled from a diverse set of trajectory patterns, including \textit{``straight"}, \textit{``curvilinear"}, \textit{``circular"}, \textit{``composite"}, etc. Camera trajectories are used to characterize the movement of the camera viewpoint, such as ``the camera slowly dollies in to capture her expression".
\noindent{(4) \textbf{Human activity}}, which covers all kinds of common human activities, including basic body movements, daily life activities, sports \& exercise, work \& study, entertainment \& leisure, and transportation \& outdoor. Each category contains a predefined set of verbs or verb phrases used for activity selection or imagination.
\noindent{(5) \textbf{Emotion}}, which includes emotion conveyed through both facial expressions and body motions. The former focuses on describing fine-grained facial expressions associated with specific emotions, for example, \textit{``A remorseful youth sighs with downcast eyes, his brows deeply creased"}. The latter expresses emotions through body movements, such as \textit{``A sad boy drags his feet on the ground, head down and arms swinging loosely"}.
\noindent{(6) \textbf{Interaction}}, which contains human self-contact, human-object interaction (HOI), human-human interaction (HHI), and human-scene interaction (HSI). It aims to evaluate the accuracy and realism of human-centric interaction behaviors in prompt-instructed video generation.
\noindent{(7) \textbf{Physical authenticity}}, which aims at evaluating the authenticity of human anatomy \& pose, motion \& physics, and object state change. It assesses whether human body structures are complete and poses are anatomically plausible, whether motion patterns obey physical laws, and whether human actions induce realistic state changes in the surrounding environment or objects.

\input{tabs/t2v_model}

Based on the above categories, we carefully design 22 category-specific prompts for ChatGPT-4o \cite{chatgpt4o} to generate prompts and extract the corresponding keywords. Please refer to the Supplementary Material for the full prompts used for ChatGPT-4o.
Specifically, ChatGPT-4o is instructed to summarize the given task and generate a detailed execution scheme \cite{peng2024dreambench++}. When ambiguities or missing requirements are detected, human experts will provide interactive feedback to revise and refine the outputs.
This iterative collaboration helps ChatGPT-4o better understand the characteristics and requirements of each category, resulting in more suitable and higher-quality prompts. All generated prompts are then manually checked and filtered. Concretely, we ensure that the prompts match the corresponding subcategories and cover both single-person and multi-person scenarios. We further verify the diversity and correctness of the prompts and the extracted keywords.
Through this internal thinking process and human-in-the-loop prompt construction pipeline, we finally obtain a set of high-quality prompts with comprehensive coverage of all subcategories.

\begin{figure*}[t]
\centering
\includegraphics[width=\linewidth]{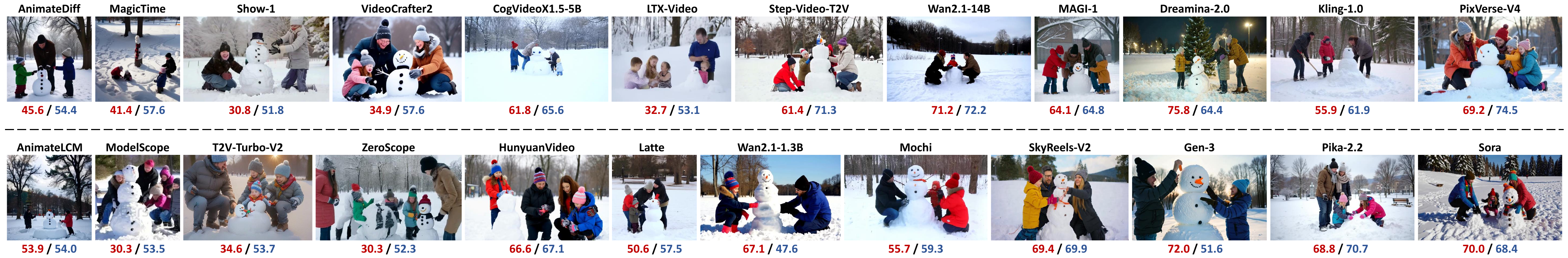}
\vspace{-6mm}
\caption{Sample video frames generated by 24 T2V models using the prompt ``Two parents and their kids are building a snowman together in a snowy park". We report \textcolor{demored}{spatial quality score} / \textcolor{demoblue}{text-video correspondence score} for each video.}
\vspace{-1mm}
\label{fig:demo1}
\end{figure*}

\begin{figure*}[t]
\centering
\includegraphics[width=\linewidth]{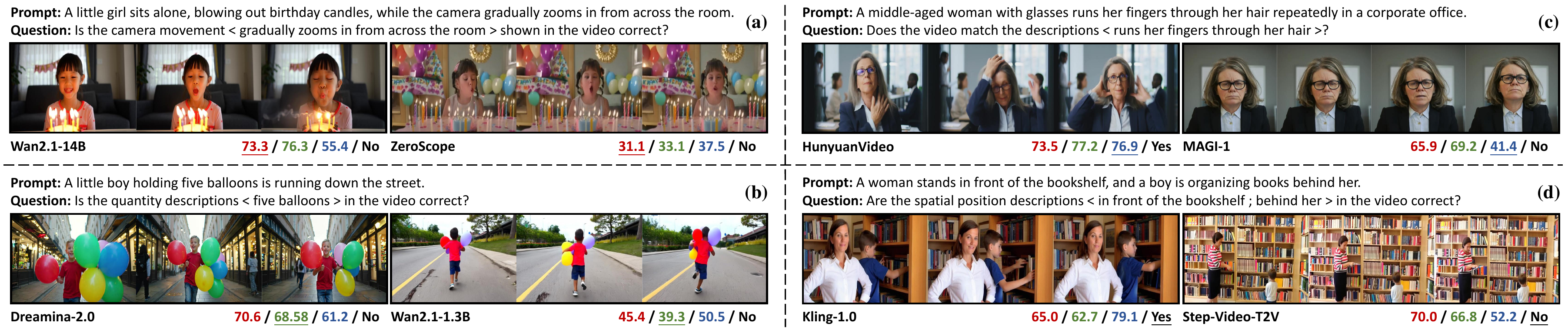}
\vspace{-6mm}
\caption{Examples from the proposed HVEval+ dataset. (a)-(d) illustrate the good and bad cases across four evaluation dimensions: spatial quality, temporal quality, text-video correspondence, and category-specific yes/no question, respectively. We report \textcolor{demored}{spatial quality score} / \textcolor{demogreen}{temporal quality score} / \textcolor{demoblue}{text-video correspondence score} / answer of the question for each video.}
\vspace{-2mm}
\label{fig:demo2}
\end{figure*}

\subsection{Video and Pair Generation}
Based on the 1,000 high-quality prompts, we utilize 24 T2V models to generate 20,000 human-centric videos, which are then used to construct 60,000 video pairs.
As shown in Table \ref{tab:t2v_model}, the T2V models cover a wide range of types, including diffusion U-Net-based models \cite{lin2024animatediff,wang2024animatelcm,yuan2025magictime,wang2023modelscope,zhang2024show,li2024t2v,chen2024videocrafter2,Zeroscope}, DiT-based models \cite{yang2024cogvideox,kong2024hunyuanvideo,hacohen2024ltx,ma2024latte,ma2025step,wan2025wan,genmo2024mochi}, autoregressive models \cite{teng2025magi,chen2025skyreels}, and commercial models \cite{Dreamina,gen3,kling,pika,PixVerse,sora}.
We use the default settings and weights for open-source models, and the official websites for commercial models to generate the videos.
Considering that absolute ratings for videos generated from different prompts and T2V models may introduce noise and ambiguity, which can lead to inconsistencies with human perceptual judgments and less reliable comparisons across models, we further enhance our dataset with pairwise data. The incorporation of pairwise data provides more accurate and human preference-aligned data for training judgment and reward models for video comparison and T2V model optimization.
Specifically, each prompt corresponds to 19 or 24 videos generated by different T2V models. We construct all possible pairs for each prompt, resulting in a total of 192,000 video pairs, from which we randomly sample 60,000 unique pairs, assigning 20,000 pairs for each evaluation dimension to construct the pairwise data of our dataset.

\subsection{Subjective Study}
To ensure the reliability of the subjective study results, we recruit 33 participants to conduct the experiments in a controlled lab setting, rather than using crowdsourcing.

As illustrated in Figure \ref{fig:dataset} (b), each video is required to be annotated with three scores across three evaluation dimensions and one yes/no question. Each pair is required to be annotated with a preference selection based on one of the three dimensions.
The three evaluation dimensions include: (1) \textbf{spatial quality}, which treats the video as a series of images and focuses on static quality aspects such as human body distortion, artifacts, clarity, color, contrast, and aesthetics; (2) \textbf{temporal quality}, which focuses on temporal consistency, motion smoothness, flicker, stutter, and judder; (3) \textbf{text-video correspondence}, which measures the degree of alignment between the generated video and the input prompt.

In the subjective study, each participant is first trained with clear rating criteria and numerous examples to fully understand the evaluation dimensions and annotation requirements. Then, a test process with 15 videos or video pairs is conducted to assess whether the participants are sufficiently trained. Only those who pass the test session are allowed to participate in the formal experiments. During the rating process, all data are randomly divided into batches for annotation, and after each batch, a data check and subject rejection procedure are carried out, excluding any subjects identified as outliers from participating in subsequent experiments.

\begin{figure}[t]
\centering
\includegraphics[width=0.95\linewidth]{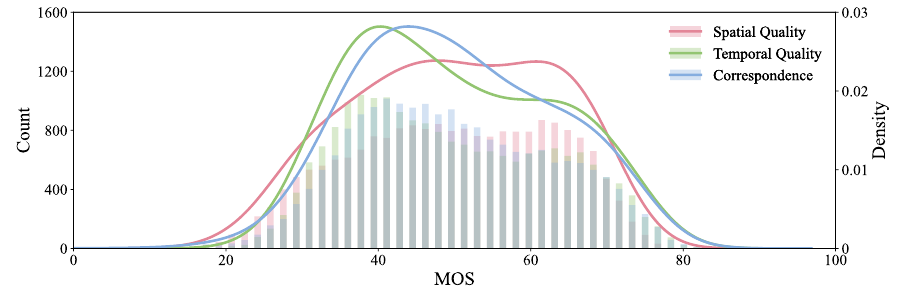}
\vspace{-2mm}
\caption{MOS distribution histograms and kernel density curves for spatial quality, temporal quality, and text-video correspondence.}
\vspace{-3mm}
\label{fig:mos}
\end{figure}

\begin{figure*}[t]
\centering
\includegraphics[width=0.99\linewidth]{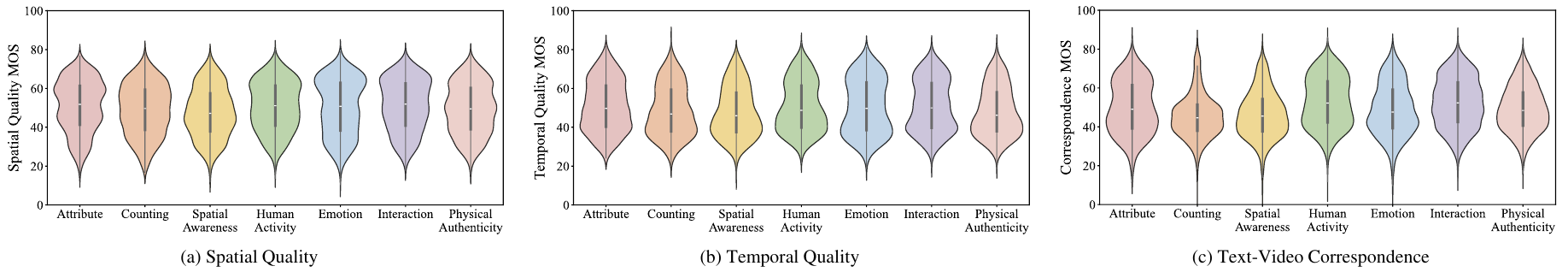}
\vspace{-3.2mm}
\caption{MOS distributions of (a) spatial quality, (b) temporal quality, and (c) text-video correspondence across the seven prompt categories.}
\vspace{-1mm}
\label{fig:violin}
\end{figure*}

\begin{figure*}[t]
\centering
\includegraphics[width=0.99\linewidth]{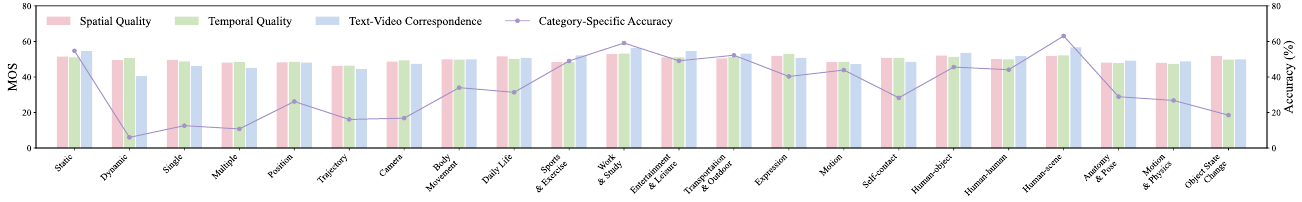}
\vspace{-3.2mm}
\caption{Comparison of average spatial quality, temporal quality, text-video correspondence, and category-specific accuracy across the 22 prompt subcategories.}
\vspace{-1mm}
\label{fig:bar_line_subcategory}
\end{figure*}

\begin{figure*}[t]
\centering
\includegraphics[width=0.99\linewidth]{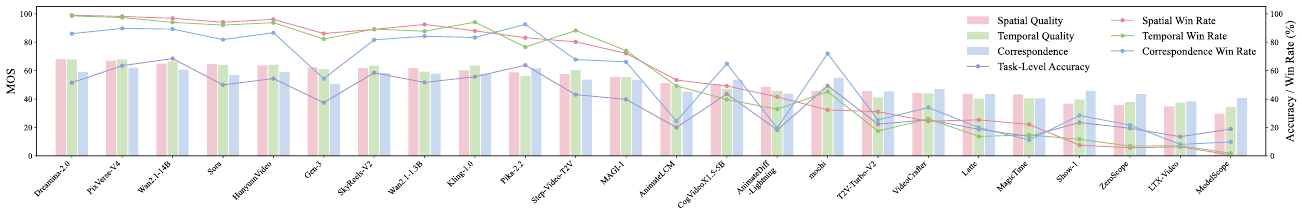}
\vspace{-3.1mm}
\caption{Comparison of the 24 T2V models in terms of average spatial quality, temporal quality, text-video correspondence, category-specific accuracy, and win rates with respect to spatial quality, temporal quality, and text–video correspondence.}
\vspace{-3mm}
\label{fig:bar_line_model}
\end{figure*}

\subsection{Subjective Data Processing}
For the three scores, we process each of them separately as follows.
We first perform outlier detection and subject rejection on all the 20,000 videos following the guidelines provided by ITU-BT.500 \cite{series2012methodology}.
The remaining valid subjective scores are then converted into Z-scores and linearly scaled to the range of $[0, 100]$ by:
\begin{equation}
  z_{ij}=\frac{r_{ij}-\mu_{i}}{\sigma_{i}}, \quad z_{ij}'=\frac{100(z_{ij}+3)}{6},
\end{equation}
where $r_{ij}$ is the raw score given by the $i$-th subject to the $j$-th video, $\mu_{i}$ and $\sigma_{i}$ are the mean score and standard deviation given by subject $i$, respectively.
Finally, the rescaled Z-scores $ z_{ij}'$ are averaged over subjects to obtain the MOSs:
\begin{equation}
  \text{MOS}_{j}=\frac{1}{N}\sum_{i=1}^{N}z_{ij}',
\end{equation}
where $j$ denotes the $j$-th video, and $N$ is the number of subjects.
For the category-specific yes/no questions and pairwise preference selections across three dimensions, the final decisions are determined by majority voting across annotators.

As a result, we obtain a total of 20k$\times$3=60k MOSs and 20k$\times$3=60k preference pairs across three dimensions (\textit{i.e.,} spatial quality, temporal quality, and text-video correspondence), and 20k category-specific question-answer pairs.

\subsection{Data Analysis}
We provide comprehensive analyses of the proposed HVEval+ dataset.
To better illustrate the proposed dataset, we first present demo frames generated by 24 T2V models in Figure \ref{fig:demo1}, and 4 pairs of good and bad examples under each of the four evaluation dimensions in Figure \ref{fig:demo2}, which demonstrate the diversity of our dataset and the distinctiveness of the annotation dimensions.
Then, the MOS distributions for spatial quality, temporal quality, and text-video correspondence are shown in Figure \ref{fig:mos}, highlighting the wide coverage and clear distinctions among the different dimensions.
We also visualize the MOS distribution for the 7 major prompt categories in Figure \ref{fig:violin}. The distributions vary across categories, underscoring the differing difficulty levels of each category.
The average MOSs and Q\&A accuracy for each subcategory, shown in Figure \ref{fig:bar_line_subcategory}, further demonstrate the distinct differences among them.
Moreover, we analyze the average MOSs, win rates, and Q\&A accuracy of videos generated by different T2V models and rank them based on spatial quality, as shown in Figure \ref{fig:bar_line_model}. It is evident that all T2V models still exhibit a noticeable gap in generating satisfactory human-centric videos, and commercial models generally perform better than open-source models.

\begin{figure*}[t]
\centering
\includegraphics[width=\linewidth]{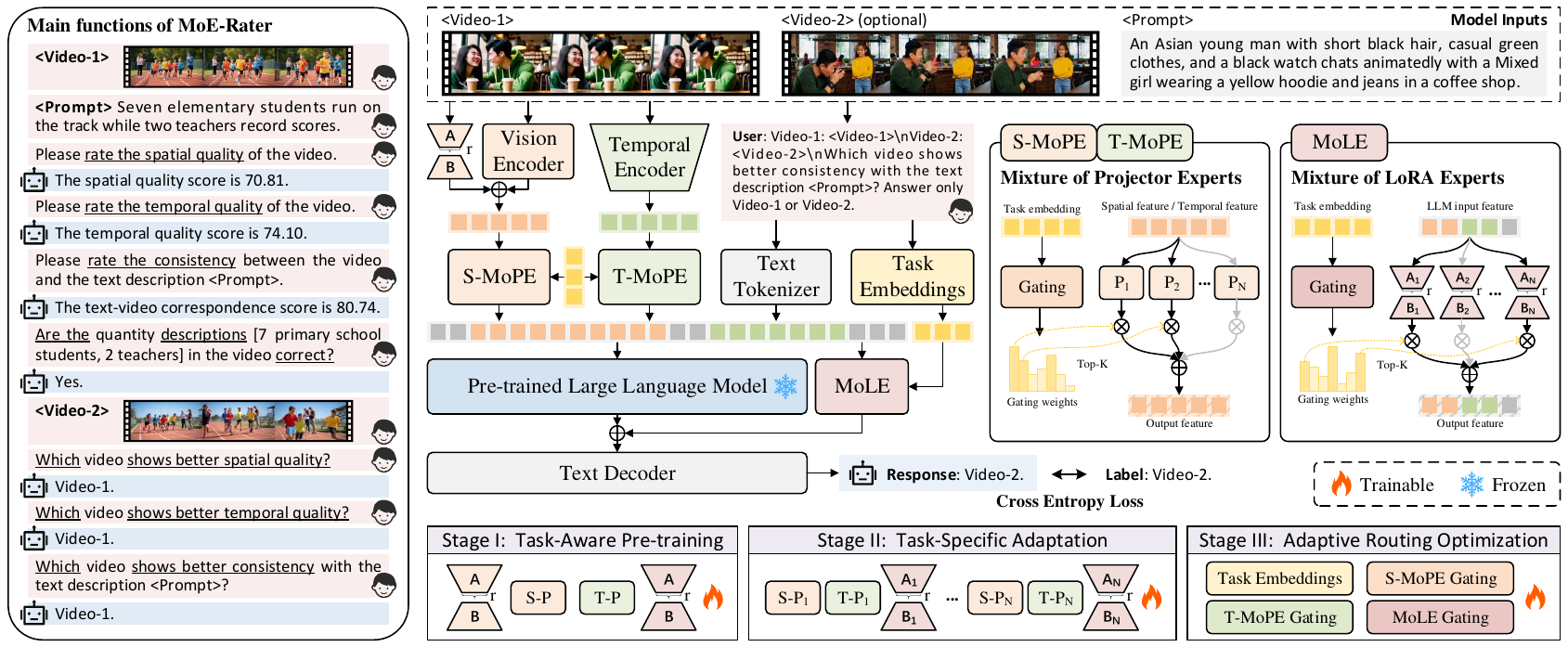}
\vspace{-5mm}
\caption{Overview of MoE-Rater. MoE-Rater serves as an all-in-one model that supports all evaluation functions, including single-video quality scoring across three evaluation dimensions, category-specific video question answering, and double-video pairwise comparison along the same dimensions. Given one or two input videos and task-specific textual instructions, MoE-Rater first extracts spatial and temporal features using the vision and temporal encoders. The extracted features are then projected into the LLM input space through S-MoPE and T-MoPE, respectively. Finally, the projected features and text embeddings are fed into a pre-trained LLM equipped with MoLE for multimodal feature fusion and reasoning, enabling multi-task responses. MoE-Rater is trained in three stages: (I) task-aware pre-training of unified projectors and LoRA modules on all tasks; (II) task-specific adaptation of projectors and LLM LoRA for each task; and (III) adaptive routing optimization that learns task embeddings and gating networks while keeping the projectors and LoRA modules frozen to activate and aggregate multiple experts.}
\vspace{-2mm}
\label{fig:method}
\end{figure*}

\input{tabs/qa_compare}

\input{tabs/mllm_compare}

\section{MoE-Rater Method}
In this section, we present our \textbf{all-in-one} method, \textbf{MoE-Rater}, which supports multi-dimensional quality rating, multi-dimensional pairwise comparison, and category-specific question answering within a single model.

\subsection{Model Overview}
An overview of MoE-Rater is illustrated in Figure \ref{fig:method}.
Given one or two input videos and task-specific textual instructions, MoE-Rater outputs corresponding textual responses for different type of question.
All tasks are formulated under a unified instruction-following paradigm, and heterogeneous single-video and double-video inputs are processed within the same multimodal framework.
Specifically, MoE-Rater first extracts spatial and temporal features using vision and temporal encoders. These features are then projected into the LLM input space through separate Mixture of Projector Experts (MoPE) modules for spatial and temporal features, respectively. The projected features, along with text embeddings, are fed into a pre-trained LLM for multimodal feature fusion and reasoning to support multi-task responses. Notably, the LLM backbone is equipped with the Mixture of LoRA Experts (MoLE) module for efficient fine-tuning and multi-task integration.

\subsection{Module Design}
\noindent\textbf{Feature Extraction.}
We extract spatial and temporal features for each input video.
The spatial features are extracted from the key frames of the input video using a vision encoder. Specifically, we uniformly sample 5 key frames $V_k$ for each input video, and use the pretrained InternViT 
\cite{zhu2025internvl3} as the vision encoder $\mathcal{E}_S$. To align the spatial features with the input space of the pre-trained LLM, we use a two-layer MLP projector $\mathcal{P}_S$ to map the spatial features into the textual embedding space, which can be formulated as:
\begin{equation}
    \boldsymbol{f}_S = \mathcal{P}_S(\mathcal{E}_S(V_k)),
\end{equation}
where $\boldsymbol{f}_S$ denotes the projected spatial features that are compatible with the LLM input space.
Since spatial features extracted from sparse frames are insufficient to capture complete temporal information, we additionally extract temporal features from continuous video frames. Specifically, we employ the pre-trained SlowFast \cite{feichtenhofer2019slowfast} as the temporal encoder $\mathcal{E}_T$. To capture both global and local temporal information, we extract features from 5 video clips (each containing 32 frames) and the full video, yielding 6 temporal features per video. A two-layer MLP projector $\mathcal{P}_T$ is also used to map the temporal features into the LLM input space:
\begin{equation}
    \boldsymbol{f}_T = \mathcal{P}_T(\mathrm{Concat}(\mathcal{E}_T(V),\mathcal{E}_T(V_{c1}),...,\mathcal{E}_T(V_{c5}))),
\end{equation}
where $\boldsymbol{f}_T$ denotes the projected temporal features compatible with the LLM input space. $\mathrm{Concat}(\cdot)$ is the concatenation operation. $V$ represents the entire input video, and $V_{ci}$ corresponds to the $i$-th video clip extracted from it.

\noindent\textbf{Mixture of Projector Experts (MoPE).}
Considering that different tasks may rely on distinct spatial and temporal clues, employing a single shared projector is suboptimal for multi-task learning. To this end, we propose two Mixture of Projector Experts (MoPE) modules for spatial and temporal features, termed as S-MoPE and T-MoPE, respectively, to better preserve task-relevant information.
Both S-MoPE and T-MoPE consist of multiple task-specific projectors and a gating network for top-k expert activation and weighted aggregation. 
Specifically, we learn a set of task embeddings to represent different tasks. The gating network is implemented as a two-layer MLP that takes the task embedding as input and predicts the routing weight for each expert. Each expert corresponds to an independent projector with the same architecture but separate parameters. The experts with the top-k largest weights are activated, while the remaining experts are masked out. The outputs of the activated experts are then aggregated through a weighted summation according to their normalized routing weights to produce the final feature representation. Taking S-MoPE as an example, its output $\boldsymbol{F}_S$ can be formulated as:
\begin{equation}
    \boldsymbol{F}_S = \sum_{i \in \mathcal{K}} w_i \boldsymbol{f}_{Si},
\end{equation}
\begin{equation}
\mathcal{K} = \mathrm{TopK}(\boldsymbol{w}, k), \quad
\boldsymbol{w} = \mathrm{Softmax}(\mathcal{G}_S(\boldsymbol{t})),
\end{equation}
where $\mathcal{K}$ denotes the set of activated top-$k$ experts. $\boldsymbol{f}_{Si}$ denotes the projected spatial features produced by $i$-th expert, and $w_i$ is the corresponding normalized gating weight. $\boldsymbol{w}=(w_1,...,w_N)$ is the gating weights vector over the $N$ experts. $\mathcal{G}_S$ denotes the gating network of S-MoPE, and $\boldsymbol{t}$ denotes the task embedding.
T-MoPE adopts the same formulation as S-MoPE with independent parameters and is applied to the temporal features.

\noindent\textbf{Mixture of LoRA Experts (MoLE).}
Similar to the feature projectors, employing a single LoRA to fine-tune the pre-trained LLM for all tasks is suboptimal in a multi-task setting. Therefore, we introduce a Mixture of LoRA Experts (MoLE) module to enhance performance across various tasks, inspired by \cite{wu2024mixture}.
Specifically, MoLE consists of multiple task-specific LoRA experts and a gating network. The gating network shares the same architecture as that in MoPE but maintains independent parameters, and takes the task embedding as input to produce the gating weights. Notably, both MoPE and MoLE use the same learnable task embeddings.
The role of the MoLE module can be formulated as follows. Given a pre-trained LLM with weight matrix $\boldsymbol{W} \in \mathbb{R}^{d\times k}$, the LoRA technique \cite{hu2022lora} adapts it by modeling the weight update with a low-rank decomposition $\Delta \boldsymbol{W}=\boldsymbol{B}\boldsymbol{A}$, where $\boldsymbol{B}\in \mathbb{R}^{d\times r}$ and $\boldsymbol{A}\in \mathbb{R}^{r\times k}$ are trainable parameter matrices, and $r\ll\{d,k\}$ for parameter efficiency. Accordingly, MoLE updates the pre-trained LLM by aggregating the trainable parameter matrices of $N$ LoRA experts $\{(\boldsymbol{B}_i,\boldsymbol{A}_i)\}_{i=1}^N$ through a weighted combination. The output $\boldsymbol{y}$ of the LLM equipped with MoLE is given by:
\begin{equation}
    \boldsymbol{y}=\boldsymbol{W}\boldsymbol{x}+
    \sum_{i \in \mathcal{K}} w_i \boldsymbol{B}_i\boldsymbol{A}_i\boldsymbol{x},
\end{equation}
\begin{equation}
\mathcal{K} = \mathrm{TopK}(\boldsymbol{w}, k), \quad
\boldsymbol{w} = \mathrm{Softmax}(\mathcal{G}_L(\boldsymbol{t})),
\end{equation}
where $\mathcal{K}$ denotes the set of activated top-$k$ experts. $\boldsymbol{w}=(w_1,...,w_N)$ is the gating weights over the $N$ experts. $\mathcal{G}_L$ is the gating network of MoLE. $\boldsymbol{t}$ represents the task embedding.

\noindent\textbf{Feature Fusion via LLM.}
The spatial and temporal features, together with text embeddings, are fed into a pre-trained LLM equipped with MoLE for multimodal feature fusion and reasoning to support all the tasks, including quality score prediction from three evaluation dimensions (\textit{i.e.}, spatial quality, temporal quality, and text-video correspondence), pairwise comparison from the same three dimensions, and category-specific yes/no question answering.
Specifically, the user prompt and feature placeholder tokens are first converted into text embeddings via a text tokenizer and an embedding layer, after which the embeddings at the placeholder positions are replaced with the extracted features $\boldsymbol{F}_S$ and $\boldsymbol{F}_T$. The resulting embeddings are then fed into both the pre-trained LLM and the MoLE module. We use the pre-trained Qwen2.5 \cite{qwen2.5} as the LLM backbone.

\subsection{Training Strategy}
MoE-Rater is trained in three stages to achieve better performance across multiple tasks.
All tasks are optimized under a unified text-generation objective with the standard cross-entropy loss.

\noindent\textbf{Stage I: Task-Aware Pre-training.}
To align heterogeneous tasks into a unified representation space and provide the MLLM with a task-aware initialization, we jointly train unified projectors $\mathcal{P}_S$ and $\mathcal{P}_T$, the vision encoder, and the LLM on all tasks, where the vision encoder and the LLM are initialized with pre-trained weights \cite{zhu2025internvl3} and each fine-tuned with their own task-shared LoRA \cite{hu2022lora} module.
In this stage, we use training data from all tasks, with each sample following the same \textit{(video, question, answer)} triplet format and each video corresponding to multiple such triplets, as illustrated in the left part of Figure \ref{fig:method}.

\noindent\textbf{Stage II: Task-Specific Adaptation.}
The unified model learned across all tasks may be suboptimal for individual tasks, as different tasks still require specialized capacity to capture task-specific reasoning patterns and achieve better performance. To this end, we fine-tune the unified spatial projector $\mathcal{P}_S$, temporal projector $\mathcal{P}_T$, and LLM LoRA $(\boldsymbol{B},\boldsymbol{A})$ learned in stage I separately for each task to obtain task-specific experts $\{\mathcal{P}_{Si}\}_{i=1}^N$, $\{\mathcal{P}_{Ti}\}_{i=1}^N$, and $\{(\boldsymbol{B}_i,\boldsymbol{A}_i)\}_{i=1}^N$. For each task, fine-tuning is conducted independently using that task’s training data while keeping the ViT LoRA frozen.

\noindent\textbf{Stage III: Adaptive Routing Optimization.}
To further leverage the complementary expertise across experts and promote mutual benefits among them, we enable dynamic activation and aggregation of multiple experts through the gating networks.
Specifically, we train the shared task embeddings, the gating network $\mathcal{G}_S$ of S-MoPE, the gating network $\mathcal{G}_T$ of T-MoPE, and the gating network $\mathcal{G}_L$ of MoLE using training data from all tasks, while keeping all projectors and LoRA modules frozen.
Decoupling routing optimization from expert learning preserves the specialized capabilities of each expert while enabling flexible expert composition at inference time.

\input{tabs/exp_human-agvqa}

\input{tabs/t2v_model_rank}

\input{tabs/ablation}

\section{Experiments}
\subsection{Implementation Details}
\noindent\textbf{Training Details of MoE-Rater.}
We use the pre-trained InternViT and Qwen2.5 from InternVL3-8B \cite{zhu2025internvl3} as our vision encoder and LLM, respectively.
The spatial projector $\mathcal{P}_S$ is initialized from the pre-trained MLP projector of InternVL3-8B \cite{zhu2025internvl3}, and the temporal projector $\mathcal{P}_T$ is randomly initialized. The task embedding dimension and the number of experts are set to 256 and 7, respectively. For both MoPE and MoLE, the top-k routing parameter k is set to 3. The task embeddings $\boldsymbol{t}_{1:N}$ and the three gating networks $\mathcal{G}_S$, $\mathcal{G}_T$, and $\mathcal{G}_L$ are also randomly initialized.
All video frames are resized to $448\times 448$ for model input.
The LoRA \cite{hu2022lora} rank is set to 8 for both the vision encoder and the LLM.
We use a learning rate of $2\times10^{-5}$, with a weight decay of $0.05$, a warm-up ratio of $0.03$, and a cosine learning rate scheduler.
For each of the three training stages, the model is trained for one epoch with a batch size of 4. The whole training process takes about 2 days on 4 NVIDIA A800 GPUs (80G).

\noindent\textbf{Evaluation Datasets.}
We conduct experiments on the proposed HVEval+ dataset and the Human-AGVQA \cite{zhang2025human} datasets.
HVEval+ is randomly split into training and test sets with a ratio of $80\%:20\%$ according to the prompt, resulting in 15,200 videos for training and 4,800 videos for testing.
Human-AGVQA \cite{zhang2025human} is randomly divided into training, validation, and test sets with a ratio of $70\%:10\%:20\%$ following its original setting, resulting in 4,200 videos for training, 600 videos for validation, and 1,200 videos for testing.

\noindent\textbf{Evaluation Metrics.}
We employ Spearman’s Rank Correlation Coefficient (SRCC) and Pearson Linear Correlation Coefficient (PLCC) to measure the consistency between predicted video quality scores and T2V model rankings with human perceptions.
For pairwise comparison tasks, we evaluate performance using better video selection accuracy.
For the category-specific question-answering task, we adopt yes/no question answering accuracy as the evaluation metric.

\subsection{Performance Evaluation}
\noindent\textbf{Evaluation of Multi-dimensional Quality Rating.}
We first evaluate the video quality score prediction performance of MoE-Rater by comparing it with VQA methods, VBench metrics, vision-language alignment metrics, and MLLMs on the proposed HVEval+ dataset.
As shown in the left part of Table \ref{tab:qa_compare}, methods trained on general-purpose video quality assessment datasets perform poorly on AI-generated human-centric videos, whereas methods trained or fine-tuned on the proposed HVEval+ dataset can achieve substantially better performance, underscoring the significance of the proposed dataset. Among all the methods, our MoE-Rater achieves the best performance across all evaluation dimensions.
We also report the performance of all MLLMs under each prompt category in Table \ref{tab:mllm_compare}, demonstrating the robustness and consistent superiority of our method.
Moreover, we evaluate our MoE-Rater on the Human-AGVQA \cite{zhang2025human} dataset, with results reported in the ``Fine-tuned" column of Table \ref{tab:exp_human-agvqa}. It is evident that our method outperforms all existing methods by a large margin.

\noindent\textbf{Evaluation of Multi-dimensional Pairwise Comparison.}
We also evaluate the double-video pairwise comparison performance of MoE-Rater across the three evaluation dimensions. For methods that cannot directly output pairwise choices, selections are determined based on the relative magnitudes of their predicted quality scores for each video in the pair. As shown in the ``Acc." columns in the left part of Table \ref{tab:qa_compare}, our method achieves the highest selection accuracy across all dimensions, demonstrating its superior and robust performance not only in score prediction but also in pairwise comparison.

\noindent\textbf{Evaluation of Category-specific Question Answering.}
As shown in the left part of Table \ref{tab:qa_compare}, pre-trained MLLMs exhibit a certain level of question-answering capability due to their strong language understanding and reasoning abilities, while fine-tuning on the proposed HVEval+ dataset can further improve their Q\&A performance. It can be observed that our MoE-Rater outperforms all other methods.

\noindent\textbf{Evaluation of T2V Model Ranking.}
An important application of VQA methods is to evaluate and compare T2V models based on the quality of their generated videos. Therefore, we further evaluate the T2V model ranking performance of our MoE-Rater by comparing it with various baseline methods. As shown in the right part of Table \ref{tab:qa_compare}, we report the SRCC between the predicted and ground-truth rankings of 24 T2V models, calculated from each model's average quality scores across three evaluation dimensions, pairwise comparison win rates across the same dimensions, and Q\&A accuracy, respectively. The results demonstrate the superior performance of our method.
Moreover, we report the average scores across various dimensions produced by both human annotations and our MoE-Rater in Table \ref{tab:t2v_compare}. We can observe that our model predicts quality scores highly consistent with human annotations, and when these scores are used to rank the T2V models, the correlation with human rankings approaches 1.0, highlighting the effectiveness of our model when applied to T2V model evaluation.

\noindent\textbf{Cross-dataset Evaluation.}
We train MoE-Rater on the proposed HVEval+ dataset, and evaluate its zero-shot performance on the Human-AGVQA \cite{zhang2025human} dataset without any additional fine-tuning or adaptation.
Although the evaluation dimensions of the two datasets are not fully identical, they are semantically related. Specifically, the predicted spatial and temporal quality scores of MoE-Rater are used as the human appearance and action continuity quality scores, respectively, while their average is used as the overall video quality score.
As shown in Table \ref{tab:exp_human-agvqa}, our model achieves the best zero-shot performance and even outperforms most of the methods fine-tuned on the Human-AGVQA \cite{zhang2025human} dataset, demonstrating the strong generalization ability of both our MoE-Rater method and the proposed HVEval+ dataset. Notably, the SRCC and PLCC of the overall score both exceed 0.75, further indicating that our dataset facilitates the learning of transferable and robust AI-generated human-centric video quality assessment capabilities and serves as a reliable benchmark.

\subsection{Ablation Study}
To validate the effectiveness of each component and training strategy of our MoE-Rater, we conduct comprehensive ablation studies on the HVEval+ dataset, as shown in Table \ref{tab:ablation}.

\noindent\textbf{Effectiveness of Temporal Feature.}
According to experiments (1) and (2), incorporating temporal features can improve both quality rating and pairwise comparison performance. The improvements are particularly significant on the temporal quality and text–video correspondence dimensions, which rely more heavily on temporal information.

\noindent\textbf{Effectiveness of LoRA Fine-tuning.}
Experiments (2)-(5) demonstrate that LoRA fine-tuning of both the ViT and the LLM improves the overall performance of our model, with ViT LoRA contributing more to spatial and temporal quality and LLM LoRA to text-video correspondence.

\noindent\textbf{Effectiveness of MoPE.}
Experiments (9) and (10) indicate that replacing the spatial and temporal projectors with two separate MoPE modules allows the model to capture task-specific features and to combine complementary information from different aspects, thereby improving overall performance.

\noindent\textbf{Effectiveness of MoLE.}
As shown in experiments (8) and (10), the introduction of the MoLE module leads to improved overall model performance. These results suggest that different tasks may exhibit distinct feature fusion and reasoning patterns, and that leveraging patterns learned from other tasks contributes to better performance.

\noindent\textbf{Effectiveness of Stage II Training.}
Based on the results of experiments (6) and (10), stage II training (\textit{i.e.}, task-specific adaptation) further improves model performance across nearly all dimensions and tasks by allowing each expert to more fully exploit its capacity through fine-tuning on task-specific data.

\noindent\textbf{Effectiveness of Stage III Training.}
The results of experiments (7) and (10) demonstrate that stage III training, \textit{i.e.}, the incorporation of the MoPE and MoLE modules for adaptive routing optimization, consistently improves the model performance across all tasks and evaluation dimensions. These results further underscore the effectiveness of aggregating multiple experts, which enables the model to leverage complementary knowledge learned by different experts and thereby enhance performance on all tasks.

\section{Limitations and Future Work}
Although HVEval+ and MoE-Rater provide a comprehensive dataset and a unified evaluation framework for AI-generated human-centric video quality assessment, several limitations remain and deserve further investigation.

From the dataset perspective, although HVEval+ covers diverse human-centric prompt categories and a wide range of representative T2V models, its coverage is still constrained by the current number of prompts and the selected generators. Therefore, the dataset may not fully cover all possible real-world human-centric generation scenarios and distortion patterns. In addition, T2V generation is evolving rapidly, and newly emerging models may exhibit different generation characteristics from those included in the current dataset. Future work could continuously expand and periodically update HVEval+ by incorporating more diverse prompts, newly released T2V models, and updated generator versions. Moreover, investigating model-agnostic distortion characteristics shared across different AI-generated videos may help reduce the influence of generator version changes and improve the robustness of quality assessment models.
Another limitation lies in the subjective annotation process. Although we follow standard subjective quality assessment procedures, human annotations are inherently labor-intensive and may still involve uncertainty due to individual differences in perception. Future research may explore semi-supervised learning strategies to better leverage large-scale unlabeled AI-generated videos, thereby improving the generalization ability of quality assessment models trained with limited human annotations.

From the model perspective, although MoE-Rater integrates spatial and temporal information within a unified MLLM-based framework, it mainly relies on generic visual and temporal representations learned by foundation models. For human-centric video quality assessment, more explicit human-related priors, such as human pose, body and facial structures, and parsing features, could be incorporated to further improve the sensitivity to human-specific distortions.
In addition, the generalization of MoE-Rater to videos generated by unseen T2V models remains an important direction for further investigation. Future work could further improve its generalization ability through large-scale pre-training on more diverse video data, leveraging unlabeled AI-generated videos, and developing more effective domain-generalization strategies.

\section{Conclusion}
In this paper, we conduct a comprehensive quality assessment study on AI-generated human-centric videos.
Specifically, we present HVEval+, the largest holistic quality assessment dataset for AI-generated human-centric videos, which consists of 1k prompts, 20k videos, and extensive human annotations, including 60k quality scores and 60k preference pairs across three dimensions (spatial quality, temporal quality, and text-video correspondence), as well as 20k category-specific Q\&A pairs.
Along with the HVEval+ dataset, we propose MoE-Rater, an MoE-inspired all-in-one method that enables multi-dimensional quality rating, multi-dimensional pairwise comparison, and category-specific question answering in a single model.
Extensive experiments on both HVEval+ and Human-AGVQA datasets demonstrate the superiority of our MoE-Rater.
We hope that the proposed dataset, benchmarks, and method will promote in-depth research on developing quality assessment methods for AI-generated videos, and further enhance the evaluation and optimization of T2V models.

% \section*{Acknowledgments}
% This should be a simple paragraph before the References to thank those individuals and institutions who have supported your work on this article.

% {\appendix[Proof of the Zonklar Equations]
% Use $\backslash${\tt{appendix}} if you have a single appendix:
% Do not use $\backslash${\tt{section}} anymore after $\backslash${\tt{appendix}}, only $\backslash${\tt{section*}}.
% If you have multiple appendixes use $\backslash${\tt{appendices}} then use $\backslash${\tt{section}} to start each appendix.
% You must declare a $\backslash${\tt{section}} before using any $\backslash${\tt{subsection}} or using $\backslash${\tt{label}} ($\backslash${\tt{appendices}} by itself
%  starts a section numbered zero.)}


% --------------------------------------------------------------------------------------------------------------
% \clearpage
\bibliographystyle{IEEEtran}
\bibliography{refs}

% --------------------------------------------------------------------------------------------------------------
% \newpage

 

% \end{IEEEbiography}



\vfill

\end{document}

%% file: tabs/dataset.tex
\begin{table*}[t]
%\vspace{-8mm}
\centering
\caption{Summary and comparison of related quality assessment datasets and evaluation benchmarks for T2V and T2I generation.}
\vspace{-0.6mm}

\resizebox{\textwidth}{!}{
\begin{tabular}{lcccccccccc}
\toprule

Dataset	&	Domain	&	Modality	&	Total Data	& Models	&	Prompts	&	Categories	&	Annotation Type	&	Annotations		&	Dimensions	&	Q\&A	\\
\midrule

% MQT	\cite{chivileva2023measuring}	&	General	&	Video	&	1,005	&	5	&	201	&	-	&	MOS	&	2,010		&	Perceptual, Correspondence	&	-	\\
EvalCrafter \cite{liu2024evalcrafter}	&	General	&	Video	&	2,500	&	5	&	700	&	17	&	MOS	&	1,024		&	Visual, Motion, Temporal, Alignment	&	-	\\
FETV \cite{liu2023fetv}	&	General	&	Video	&	2,476	&	4	&	619	&	5	&	MOS	&	7,428		&	Static, Temporal, Alignment	&	-	\\
VBench \cite{huang2024vbench}	&	General	&	Video	&	-	&	4	&	800	&	16	&	Pair	&	-		&	Overall	&	-	\\
GenAI-Bench \cite{li2024evaluating}	&	General	&	Video	&	-	&	4	&	800	&	8	&	MOS	&	-		&	Correspondence	&	-	\\
LGVQ \cite{zhang2024benchmarking}	&	General	&	Video	&	2,808	&	6	&	468	&	-	&	MOS	&	8,424		&	Spatial, Temporal, Alignment	&	-	\\
T2VQA-DB \cite{kou2024subjective}	&	General	&	Video	&	10,000	&	9	&	1,000	&	-	&	MOS	&	10,000		&	Overall	&	-	\\
T2V-CompBench \cite{sun2024t2v}	&	General	&	Video	&	651	&	23	&	1,400	&	7	&	MOS	&	651		&	Correspondence	&	\checkmark	\\
Vbench-2.0 \cite{zheng2025vbench}	&	General	&	Video	&	-	&	4	&	1,260	&	18	&	Pair	&	-		&	Correspondence	&	\checkmark	\\
% \hdashline
\midrule

FaceQ-Gen \cite{liu2025f}	&	Face	&	Image	&	4,032	&	14	&	288	&	9	&	MOS	&	12,096		&	Quality, Authenticity, Correspondence	&	-	\\
FineTFIQA \cite{gao2025multi}	&	Face	&	Image	&	7,218	&	8	&	1,000	&	12	&	MOS	&	28,872		&	Perceptual, Human Likeness, Attractiveness, Consistency	&	-	\\
AGHI-QA \cite{li2025aghi}	&	Human	&	Image	&	4,000	&	10	&	400	&	7	&	MOS	&	8,000		&	Perceptual, Correspondence	&	-	\\
% \hdashline
\midrule

Human-AGVQA	\cite{zhang2025human} &	Human	&	Video	&	6,000	&	15	&	400	&	8	&	MOS	&	18,000		&	Appearance, Continuity, Overall	&	-	\\
HVEval \cite{wu2025hveval}	&	Human	&	Video	&	20,000	&	24	&	1,000	&	22	&	MOS	&	80,000		&	Spatial, Temporal, Correspondence	&	\checkmark	\\

\rowcolor[gray]{.92}
\textbf{HVEval+ (Ours)}	&	Human	&	Video	&	20,000	&	24	&	1,000	&	22	&	MOS \& Pair	&	140,000		&	Spatial, Temporal, Correspondence	&	\checkmark	\\

\bottomrule
\end{tabular}
\label{tab:dataset}
}
\vspace{-2mm}
\centering
\end{table*}

%% file: tabs/t2v_model.tex
\begin{table}[t]
%\vspace{-8mm}
\centering
\caption{Overview of T2V models used to construct HVEval+ dataset.}
\vspace{-0.5mm}

\resizebox{0.97\linewidth}{!}{
\begin{tabular}{lccccc}
\toprule

Model	&	Year	&	Resolution	&	Frames	&	FPS	&	Type	\\
\midrule

AnimateDiff \cite{lin2024animatediff}	&	2025	&	512×512	&	16	&	8	&	Diffusion U-Net	\\
AnimateLCM \cite{wang2024animatelcm}	&	2024	&	512×512	&	16	&	8	&	Diffusion U-Net	\\
MagicTime \cite{yuan2025magictime}	&	2024	&	512×512	&	16	&	8	&	Diffusion U-Net	\\
ModelScope \cite{wang2023modelscope}	&	2023	&	256×256	&	16	&	10	&	Diffusion U-Net	\\
Show-1 \cite{zhang2024show}	&	2023	&	576×320	&	29	&	8	&	Diffusion U-Net	\\
T2V-Turbo-V2 \cite{li2024t2v}	&	2024	&	512×320	&	16	&	8	&	Diffusion U-Net	\\
VideoCrafter \cite{chen2024videocrafter2}	&	2024	&	512×320	&	16	&	10	&	Diffusion U-Net	\\
ZeroScope \cite{Zeroscope}	&	2023	&	576×320	&	24	&	10	&	Diffusion U-Net	\\
% \hdashline
\midrule

CogVideoX1.5-5B \cite{yang2024cogvideox}	&	2025	&	1360×768	&	81	&	8	&	Diffusion Transformer	\\
HunyuanVideo \cite{kong2024hunyuanvideo}	&	2025	&	1280×720	&	129	&	24	&	Diffusion Transformer	\\
LTX-Video \cite{hacohen2024ltx}	&	2025	&	704×480	&	161	&	24	&	Diffusion Transformer	\\
Latte \cite{ma2024latte}	&	2024	&	512×512	&	16	&	8	&	Diffusion Transformer	\\
Step-Video-T2V \cite{ma2025step}	&	2025	&	992×544	&	51	&	25	&	Diffusion Transformer	\\
Wan2.1-1.3B \cite{wan2025wan}	&	2025	&	832×480	&	81	&	16	&	Diffusion Transformer	\\
Wan2.1-14B \cite{wan2025wan}	&	2025	&	1280×720	&	81	&	16	&	Diffusion Transformer	\\
Mochi \cite{genmo2024mochi}	&	2024	&	848×480	&	79	&	30	&	Diffusion Transformer	\\
% \hdashline
\midrule

MAGI-1 \cite{teng2025magi}	&	2025	&	720×720	&	96	&	24	&	Autoregressive	\\
SkyReels-V2 \cite{chen2025skyreels}	&	2025	&	1280×720	&	97	&	24	&	Autoregressive	\\
% \hdashline
\midrule

Dreamina-2.0 \cite{Dreamina}	&	2024	&	1472×832	&	121	&	24	&	Commercial	\\
Gen-3 \cite{gen3}	&	2024	&	1280×768	&	128	&	24	&	Commercial	\\
Kling-1.0 \cite{kling}	&	2024	&	1280×720	&	153	&	30	&	Commercial	\\
Pika-2.2 \cite{pika}	&	2025	&	1280×720	&	121	&	24	&	Commercial	\\
PixVerse-V4	\cite{PixVerse}	&	2025	&	1280×720	&	161	&	30	&	Commercial	\\
Sora \cite{sora}	&	2025	&	1280×720	&	150	&	30	&	Commercial	\\

\bottomrule
\end{tabular}
\label{tab:t2v_model}
}
\vspace{-2.5mm}
\centering
\end{table}

%% file: tabs/qa_compare.tex
\begin{table*}[t]
\centering
% \belowrulesep=0pt
% \aboverulesep=0pt
% \renewcommand\arraystretch{0.9}
\caption{Comparison with state-of-the-art methods on the HVEval+ dataset in terms of video quality assessment and T2V model ranking. $\diamondsuit$, $\spadesuit$, $\heartsuit$, and $\clubsuit$ denote video quality assessment methods, VBench \cite{huang2024vbench} metrics, vision-language alignment metrics, and multimodal large language models, respectively. The $\rho$ and $r$ below the ``Video Quality Assessment" task denote SRCC and PLCC based on per-video scores, while the $\rho$ and $\rho_p$ below the ``T2V Model Ranking" task denote SRCC calculated from per-model average scores or Q\&A accuracy, and model win rates, respectively. The abbreviation ``Acc." below ``Q\&A" represents category-specific question-answering accuracy (\%), while in all other columns it denotes pairwise comparison accuracy (\%). * and \dag\, indicate models trained and fine-tuned on the HVEval+ dataset, respectively. The best and runner-up performances are bold and underlined, respectively.}
\vspace{-0.5mm}

\resizebox{\linewidth}{!}{
\begin{tabular}{l||ccc|ccc|ccc|c||cc|cc|cc|c}
\toprule
Task  & \multicolumn{10}{c||}{\textbf{Video Quality Assessment}}  & \multicolumn{7}{c}{\textbf{T2V Model Ranking}} \\
\midrule

Dimension  & \multicolumn{3}{c|}{Spatial Quality}  & \multicolumn{3}{c|}{Temporal Quality}  & \multicolumn{3}{c|}{Correspondence}  & \multicolumn{1}{c||}{Q\&A}  & \multicolumn{2}{c|}{Spatial Quality}  & \multicolumn{2}{c|}{Temporal Quality}  & \multicolumn{2}{c|}{Correspondence}  & \multicolumn{1}{c}{Q\&A} \\
\midrule

% Method / Metric	&	SRCC$\uparrow$	&	PLCC$\uparrow$	&	Acc.$\uparrow$  &	SRCC$\uparrow$	&	PLCC$\uparrow$	&	Acc.$\uparrow$  &	SRCC$\uparrow$	&	PLCC$\uparrow$	&	Acc.$\uparrow$	&	Acc.$\uparrow$	&	SRCC$\uparrow$	&	SRCC$\uparrow$	&	SRCC$\uparrow$	&	SRCC$\uparrow$	&	SRCC$\uparrow$	&	SRCC$\uparrow$	&	SRCC$\uparrow$ \\
% \midrule

Method / Metric	&	$\rho$\,$\uparrow$	&	$r$\,$\uparrow$	&	Acc.\,$\uparrow$  &	$\rho$\,$\uparrow$	&	$r$\,$\uparrow$	&	Acc.\,$\uparrow$  &	$\rho$\,$\uparrow$	&	$r$\,$\uparrow$	&	Acc.\,$\uparrow$	&	Acc.\,$\uparrow$	&	$\rho$\,$\uparrow$	&	$\rho_p$\,$\uparrow$	&	$\rho$\,$\uparrow$	&	$\rho_p$\,$\uparrow$	&	$\rho$\,$\uparrow$	&	$\rho_p$\,$\uparrow$	&	$\rho$\,$\uparrow$ \\
\midrule

$\diamondsuit$ GSTVQA \cite{chen2021learning}	&	0.324 	&	0.315 	&	71.38 	&	0.309 	&	0.299 	&	70.13 	&	0.180 	&	0.177 	&	64.52 	&	52.23 	&	0.710 	&	0.813 	&	0.687 	&	0.827 	&	0.593 	&	0.673 	&	0.545 	\\
$\diamondsuit$ SimpleVQA \cite{sun2022deep}	&	0.709 	&	0.703 	&	87.40 	&	0.628 	&	0.621 	&	84.48 	&	0.412 	&	0.409 	&	73.35 	&	58.58 	&	0.925 	&	0.965 	&	0.876 	&	0.950 	&	0.786 	&	0.825 	&	0.770 	\\
$\diamondsuit$ MinimalisticVQA \cite{sun2024analysis}	&	0.703 	&	0.710 	&	87.48 	&	0.619 	&	0.618 	&	83.94 	&	0.396 	&	0.395 	&	72.94 	&	57.88 	&	0.929 	&	0.953 	&	0.888 	&	0.919 	&	0.746 	&	0.777 	&	0.689 	\\
$\diamondsuit$ FastVQA \cite{wu2022fast}	&	0.666 	&	0.608 	&	86.25 	&	0.584 	&	0.509 	&	81.31 	&	0.402 	&	0.352 	&	72.94 	&	48.35 	&	0.905 	&	0.965 	&	0.849 	&	0.929 	&	0.784 	&	0.823 	&	0.661 	\\
$\diamondsuit$ Dover \cite{wu2023exploring}	&	0.602 	&	0.612 	&	83.46 	&	0.516 	&	0.518 	&	78.58 	&	0.351 	&	0.352 	&	70.42 	&	55.58 	&	0.854 	&	0.936 	&	0.787 	&	0.933 	&	0.722 	&	0.777 	&	0.629 	\\
$\diamondsuit$ KSVQE \cite{lu2024kvq}	&	0.228 	&	0.216 	&	60.81 	&	0.229 	&	0.218 	&	61.46 	&	0.181 	&	0.155 	&	60.73 	&	57.35 	&	0.360 	&	0.527 	&	0.336 	&	0.531 	&	0.369 	&	0.582 	&	0.492 	\\
$\diamondsuit$ Q-Align \cite{wu2023q}	&	0.767 	&	0.747 	&	90.44 	&	0.681 	&	0.642 	&	86.38 	&	0.469 	&	0.447 	&	78.35 	&	55.83 	&	0.923 	&	0.966 	&	0.873 	&	0.961 	&	0.806 	&	0.871 	&	0.766 	\\
\midrule

$\spadesuit$ Imaging Quality \cite{huang2024vbench}	&	0.360 	&	0.383 	&	68.94 	&	0.314 	&	0.318 	&	67.92 	&	0.195 	&	0.224 	&	60.88 	&	49.46 	&	0.617 	&	0.810 	&	0.618 	&	0.810 	&	0.456 	&	0.466 	&	0.565 	\\
$\spadesuit$ Aesthetic Quality \cite{huang2024vbench}	&	0.406 	&	0.425 	&	73.25 	&	0.383 	&	0.390 	&	71.50 	&	0.279 	&	0.298 	&	65.40 	&	54.58 	&	0.763 	&	0.877 	&	0.694 	&	0.864 	&	0.569 	&	0.632 	&	0.515 	\\
$\spadesuit$ Motion Smoothness \cite{huang2024vbench}	&	0.330 	&	0.310 	&	66.92 	&	0.395 	&	0.341 	&	70.15 	&	0.222 	&	0.204 	&	62.13 	&	42.15 	&	0.483 	&	0.543 	&	0.550 	&	0.621 	&	0.430 	&	0.513 	&	0.344 	\\
$\spadesuit$ Subject Consistency \cite{huang2024vbench}	&	0.079 	&	0.133 	&	51.92 	&	0.144 	&	0.165 	&	54.65 	&	0.045 	&	0.084 	&	48.10 	&	44.77 	&	0.137 	&	0.103 	&	0.205 	&	0.214 	&	0.041 	&	-0.031 	&	0.032 	\\
$\spadesuit$ Overall Consistency \cite{huang2024vbench}	&	-0.033 	&	-0.042 	&	51.46 	&	-0.030 	&	-0.034 	&	52.00 	&	-0.041 	&	-0.051 	&	50.10 	&	50.21 	&	0.171 	&	0.230 	&	0.173 	&	0.199 	&	-0.010 	&	0.014 	&	-0.044 	\\
\midrule

$\heartsuit$ CLIPScore \cite{hessel2021clipscore}	&	0.040 	&	0.057 	&	50.40 	&	0.018 	&	0.026 	&	51.38 	&	0.243 	&	0.259 	&	60.31 	&	54.52 	&	-0.128 	&	-0.059 	&	-0.102 	&	-0.017 	&	-0.032 	&	0.404 	&	0.089 	\\
$\heartsuit$ BLIPScore \cite{li2022blip}	&	0.110 	&	0.116 	&	55.31 	&	0.100 	&	0.102 	&	54.98 	&	0.335 	&	0.343 	&	67.35 	&	54.90 	&	0.202 	&	0.374 	&	0.170 	&	0.292 	&	0.407 	&	0.712 	&	0.366 	\\
$\heartsuit$ AestheticScore \cite{schuhmann2022laion}	&	0.493 	&	0.511 	&	74.73 	&	0.446 	&	0.450 	&	73.46 	&	0.298 	&	0.313 	&	66.98 	&	54.31 	&	0.643 	&	0.815 	&	0.615 	&	0.803 	&	0.448 	&	0.605 	&	0.415 	\\
$\heartsuit$ PickScore \cite{kirstain2023pick}	&	0.369 	&	0.376 	&	73.27 	&	0.349 	&	0.354 	&	72.94 	&	0.430 	&	0.442 	&	74.10 	&	60.79 	&	0.667 	&	0.823 	&	0.650 	&	0.870 	&	0.646 	&	0.763 	&	0.628 	\\
$\heartsuit$ ImageReward \cite{xu2023imagereward}	&	0.271 	&	0.283 	&	64.38 	&	0.267 	&	0.274 	&	63.92 	&	0.456 	&	0.474 	&	72.79 	&	59.58 	&	0.536 	&	0.679 	&	0.523 	&	0.690 	&	0.623 	&	0.793 	&	0.685 	\\
$\heartsuit$ VQAScore \cite{lin2024evaluating}	&	0.237 	&	0.214 	&	60.98 	&	0.231 	&	0.205 	&	60.98 	&	0.525 	&	0.502 	&	75.23 	&	56.04 	&	0.550 	&	0.746 	&	0.558 	&	0.767 	&	0.831 	&	0.912 	&	0.777 	\\
\midrule

$\clubsuit$ Video-ChatGPT (7B) \cite{maaz2023video}	&	-0.071 	&	-0.039 	&	47.31 	&	-0.080 	&	-0.069 	&	45.81 	&	0.015 	&	-0.005 	&	54.21 	&	60.85 	&	-0.378 	&	-0.260 	&	-0.389 	&	-0.229 	&	-0.120 	&	0.442 	&	0.347 	\\
$\clubsuit$ VideoLLaMA2 (7B) \cite{cheng2024videollama}	&	-0.069 	&	-0.037 	&	46.29 	&	-0.048 	&	-0.048 	&	49.46 	&	0.433 	&	0.344 	&	69.85 	&	60.90 	&	-0.035 	&	-0.157 	&	-0.112 	&	0.024 	&	0.823 	&	0.764 	&	-0.075 	\\
$\clubsuit$ VideoLLaMA3 (7B) \cite{zhang2025videollama}	&	0.175 	&	0.219 	&	57.69 	&	0.095 	&	0.074 	&	55.33 	&	0.363 	&	0.389 	&	66.40 	&	64.10 	&	0.692 	&	0.576 	&	0.632 	&	0.330 	&	0.851 	&	0.894 	&	0.811 	\\
$\clubsuit$ Qwen2-VL (7B) \cite{Qwen2-VL}	&	0.429 	&	0.397 	&	65.02 	&	0.393 	&	0.333 	&	69.75 	&	0.405 	&	0.420 	&	67.63 	&	63.75 	&	0.885 	&	0.550 	&	0.864 	&	0.812 	&	0.779 	&	0.715 	&	0.286 	\\
$\clubsuit$ Qwen2.5-VL (7B) \cite{Qwen2.5-VL}	&	0.419 	&	0.387 	&	62.44 	&	0.307 	&	0.253 	&	63.27 	&	0.506 	&	0.439 	&	69.83 	&	63.23 	&	0.855 	&	0.456 	&	0.835 	&	0.579 	&	0.841 	&	0.747 	&	0.899 	\\
$\clubsuit$ DeepSeek-VL (7B) \cite{lu2024deepseek}	&	0.098 	&	-0.011 	&	53.33 	&	0.123 	&	-0.020 	&	54.92 	&	0.008 	&	-0.007 	&	50.54 	&	60.17 	&	0.290 	&	0.341 	&	0.318 	&	0.418 	&	-0.269 	&	0.225 	&	-0.627 	\\
$\clubsuit$ UI-TARS (7B) \cite{qin2025ui}	&	0.495 	&	0.400 	&	68.79 	&	0.351 	&	0.274 	&	64.88 	&	0.405 	&	0.365 	&	70.73 	&	68.83 	&	0.858 	&	0.677 	&	0.702 	&	0.756 	&	0.922 	&	0.815 	&	0.899 	\\
$\clubsuit$ InternVL2.5 (8B) \cite{chen2024expanding}	&	0.423 	&	0.338 	&	70.44 	&	0.323 	&	0.321 	&	64.83 	&	0.537 	&	0.472 	&	74.02 	&	60.83 	&	0.804 	&	0.784 	&	0.527 	&	0.576 	&	0.848 	&	0.894 	&	-0.453 	\\
$\clubsuit$ InternVL3 (8B) \cite{zhu2025internvl3}	&	0.268 	&	0.241 	&	58.08 	&	0.028 	&	0.060 	&	47.50 	&	0.442 	&	0.385 	&	69.40 	&	58.31 	&	0.341 	&	0.293 	&	-0.123 	&	-0.117 	&	0.825 	&	0.790 	&	0.841 	\\
% $\clubsuit$ CogVLM2-Video (12B) \cite{hong2024cogvlm2}	&	-0.044 	&	-0.044 	&	46.69 	&	0.023 	&	0.023 	&	49.69 	&	0.172 	&	0.190 	&	58.35 	&	53.56 	&	-0.563 	&	-0.058 	&	0.130 	&	0.024 	&	0.655 	&	0.358 	&	0.656 	\\
$\clubsuit$ InternVL3 (14B) \cite{zhu2025internvl3}	&	0.232 	&	0.268 	&	55.88 	&	0.261 	&	0.215 	&	59.13 	&	0.547 	&	0.509 	&	73.42 	&	68.77 	&	0.428 	&	0.263 	&	0.493 	&	0.336 	&	0.792 	&	0.835 	&	0.791 	\\
$\clubsuit$ CogAgent (18B) \cite{hong2024cogagent}	&	0.161 	&	0.228 	&	60.69 	&	0.254 	&	0.245 	&	66.42 	&	0.176 	&	0.221 	&	58.60 	&	50.88 	&	0.591 	&	0.577 	&	0.835 	&	0.770 	&	0.675 	&	0.378 	&	-0.379 	\\
$\clubsuit$ InternVL3 (38B) \cite{zhu2025internvl3}	&	0.269 	&	0.289 	&	56.63 	&	0.182 	&	0.209 	&	54.73 	&	0.622 	&	0.550 	&	79.90 	&	63.25 	&	0.323 	&	0.223 	&	0.164 	&	0.169 	&	0.847 	&	0.937 	&	0.176 	\\
$\clubsuit$ Qwen2.5-VL (72B) \cite{Qwen2.5-VL}	&	0.498 	&	0.440 	&	70.71 	&	0.284 	&	0.151 	&	60.52 	&	0.558 	&	0.481 	&	73.94 	&	67.65 	&	0.884 	&	0.679 	&	0.642 	&	0.465 	&	0.797 	&	0.792 	&	0.835 	\\
$\clubsuit$ UI-TARS (72B) \cite{qin2025ui}	&	0.618 	&	0.617 	&	80.56 	&	0.581 	&	0.541 	&	81.71 	&	0.501 	&	0.426 	&	72.52 	&	72.65 	&	0.890 	&	0.956 	&	0.874 	&	0.941 	&	0.828 	&	0.938 	&	0.873 	\\
$\clubsuit$ InternVL3 (78B) \cite{zhu2025internvl3}	&	0.450 	&	0.366 	&	67.77 	&	0.417 	&	0.371 	&	68.77 	&	0.612 	&	0.490 	&	80.02 	&	60.83 	&	0.796 	&	0.640 	&	0.711 	&	0.731 	&	0.820 	&	0.937 	&	-0.283 	\\
\midrule

$\diamondsuit$ SimpleVQA \cite{sun2022deep}\,*	&	0.842 	&	0.843 	&	94.35 	&	0.804 	&	0.782 	&	93.04 	&	0.549 	&	0.527 	&	79.48 	&	62.17 	&	0.976 	&	\underline{0.994} 	&	0.974 	&	\textbf{0.991} 	&	0.863 	&	0.872 	&	0.781 	\\
$\diamondsuit$ FastVQA \cite{wu2022fast}\,*	&	0.828 	&	0.831 	&	93.27 	&	0.799 	&	0.793 	&	92.71 	&	0.540 	&	0.542 	&	80.44 	&	61.46 	&	0.962 	&	0.985 	&	0.951 	&	0.978 	&	0.892 	&	0.913 	&	0.802 	\\
$\diamondsuit$ Dover \cite{wu2023exploring}\,*	&	0.832 	&	0.834 	&	94.50 	&	0.794 	&	0.793 	&	92.25 	&	0.540 	&	0.542 	&	80.31 	&	63.35 	&	\underline{0.982} 	&	\underline{0.994} 	&	0.965 	&	0.986 	&	0.951 	&	0.931 	&	0.834 	\\
$\clubsuit$ Qwen2.5-VL (7B) \cite{Qwen2.5-VL}\,\dag	&	\underline{0.860} 	&	\underline{0.869} 	&	\underline{95.21} 	&	\underline{0.816} 	&	\underline{0.821} 	&	\underline{93.08} 	&	\underline{0.755} 	&	0.739 	&	88.40 	&	78.33 	&	0.977 	&	0.989 	&	\textbf{0.977} 	&	0.986 	&	\underline{0.982} 	&	\textbf{0.979} 	&	\underline{0.956} 	\\
% $\clubsuit$ UI-TARS (7B) \cite{qin2025ui}\,\dag	&	0.859 	&	0.862 	&	95.48 	&	0.824 	&	0.827 	&	93.46 	&	0.749 	&	0.740 	&	88.42 	&	79.81 	&	0.983 	&	0.997 	&	0.983 	&	0.989 	&	0.982 	&	0.973 	&	0.939 	\\
$\clubsuit$ InternVL2.5 (8B) \cite{chen2024expanding}\,\dag	&	0.839 	&	0.846 	&	93.60 	&	0.785 	&	0.785 	&	91.75 	&	0.751 	&	\underline{0.746} 	&	\underline{88.77} 	&	\underline{80.71} 	&	0.976 	&	0.992 	&	0.964 	&	\underline{0.990} 	&	0.962 	&	\underline{0.975} 	&	0.937 	\\
$\clubsuit$ InternVL3 (8B) \cite{zhu2025internvl3}\,\dag	&	0.855 	&	0.862 	&	94.83 	&	0.808 	&	0.804 	&	92.96 	&	0.692 	&	0.616 	&	86.83 	&	79.56 	&	0.977 	&	\textbf{0.997} 	&	0.963 	&	\underline{0.990} 	&	0.970 	&	\underline{0.975} 	&	0.941 	\\

\rowcolor[gray]{.92}
\textbf{MoE-Rater (Ours)}\,*	&	\textbf{0.891} 	&	\textbf{0.897} 	&	\textbf{96.56} 	&	\textbf{0.868} 	&	\textbf{0.870} 	&	\textbf{95.63} 	&	\textbf{0.787} 	&	\textbf{0.779} 	&	\textbf{90.54} 	&	\textbf{81.48} 	&	\textbf{0.991} 	&	\textbf{0.997} 	&	\underline{0.975} 	&	0.980 	&	\textbf{0.984} 	&	\underline{0.975} 	&	\textbf{0.974} 	\\

\bottomrule
\end{tabular}
}
\vspace{-2mm}
\label{tab:qa_compare}
\end{table*}
% \vspace{-12pt}

%% file: tabs/mllm_compare.tex
\begin{table*}[t]
\centering
\caption{Comparison with state-of-the-art multimodal large language models across different prompt categories. $\rho_s$, $\rho_t$, $\rho_c$, and ``Acc." denote SRCC based on three dimensions (spatial quality, temporal quality, and text-video correspondence) and category-specific question-answering accuracy (\%), respectively. * and \dag\, indicate models trained and fine-tuned on the HVEval+ dataset, respectively. The best and runner-up performances are bold and underlined, respectively.}
\vspace{-0.5mm}

\resizebox{0.9\linewidth}{!}{
\begin{tabular}{l||cccc|cccc|cccc|cccc}
\toprule

Prompt Category  & \multicolumn{4}{c|}{\textbf{Attribute}}  & \multicolumn{4}{c|}{\textbf{Counting}}  & \multicolumn{4}{c|}{\textbf{Spatial Awareness}}  & \multicolumn{4}{c}{\textbf{Human Activity}} \\
\midrule
Method / Metric	&	$\rho_s$\,$\uparrow$	&	$\rho_t$\,$\uparrow$	&	$\rho_c$\,$\uparrow$  &	Acc.\,$\uparrow$	&	$\rho_s$\,$\uparrow$	&	$\rho_t$\,$\uparrow$	&	$\rho_c$\,$\uparrow$  &	Acc.\,$\uparrow$  &	$\rho_s$\,$\uparrow$	&	$\rho_t$\,$\uparrow$	&	$\rho_c$\,$\uparrow$  &	Acc.\,$\uparrow$  &	$\rho_s$\,$\uparrow$	&	$\rho_t$\,$\uparrow$	&	$\rho_c$\,$\uparrow$  &	Acc.\,$\uparrow$ \\
\midrule
Video-ChatGPT (7B) \cite{maaz2023video}	&	-0.047 	&	-0.095 	&	-0.080 	&	65.05 	&	-0.104 	&	-0.191 	&	-0.079 	&	83.80 	&	-0.097 	&	-0.062 	&	0.023 	&	81.25 	&	-0.075 	&	-0.084 	&	0.053 	&	50.09 	\\
VideoLLaMA2 (7B) \cite{cheng2024videollama}	&	0.025 	&	-0.046 	&	0.577 	&	65.05 	&	0.015 	&	-0.081 	&	0.418 	&	84.03 	&	-0.108 	&	-0.027 	&	0.389 	&	81.41 	&	-0.042 	&	-0.026 	&	0.442 	&	50.09 	\\
VideoLLaMA3 (7B) \cite{zhang2025videollama}	&	0.061 	&	0.093 	&	0.352 	&	65.51 	&	0.176 	&	0.166 	&	0.367 	&	84.03 	&	0.151 	&	0.084 	&	0.378 	&	82.05 	&	0.184 	&	0.071 	&	0.358 	&	59.90 	\\
Qwen2-VL (7B) \cite{Qwen2-VL}	&	0.444 	&	0.433 	&	0.501 	&	66.67 	&	0.479 	&	0.397 	&	0.266 	&	84.72 	&	0.437 	&	0.405 	&	0.347 	&	81.09 	&	0.345 	&	0.364 	&	0.407 	&	60.68 	\\
Qwen2.5-VL (7B) \cite{Qwen2.5-VL}	&	0.383 	&	0.335 	&	0.578 	&	69.68 	&	0.524 	&	0.379 	&	0.428 	&	71.76 	&	0.391 	&	0.279 	&	0.465 	&	59.29 	&	0.378 	&	0.303 	&	0.537 	&	69.01 	\\
DeepSeek-VL (7B) \cite{lu2024deepseek}	&	0.169 	&	0.158 	&	0.089 	&	62.96 	&	0.177 	&	0.176 	&	0.043 	&	84.03 	&	0.093 	&	0.151 	&	-0.015 	&	80.61 	&	0.057 	&	0.090 	&	-0.044 	&	48.78 	\\
UI-TARS (7B) \cite{qin2025ui}	&	0.534 	&	0.318 	&	0.487 	&	76.62 	&	0.542 	&	0.437 	&	0.258 	&	89.12 	&	0.535 	&	0.344 	&	0.449 	&	79.65 	&	0.486 	&	0.341 	&	0.405 	&	60.68 	\\
InternVL2.5 (8B) \cite{chen2024expanding}	&	0.423 	&	0.242 	&	0.605 	&	65.05 	&	0.487 	&	0.339 	&	0.570 	&	83.80 	&	0.382 	&	0.370 	&	0.459 	&	81.41 	&	0.442 	&	0.319 	&	0.564 	&	50.00 	\\
InternVL3 (8B) \cite{zhu2025internvl3}	&	0.209 	&	-0.094 	&	0.542 	&	66.20 	&	0.285 	&	-0.021 	&	0.494 	&	41.44 	&	0.257 	&	0.091 	&	0.360 	&	48.88 	&	0.213 	&	0.045 	&	0.453 	&	65.28 	\\
InternVL3 (14B) \cite{zhu2025internvl3}	&	0.239 	&	0.134 	&	0.659 	&	75.23 	&	0.321 	&	0.304 	&	0.532 	&	79.40 	&	0.222 	&	0.301 	&	0.473 	&	65.54 	&	0.185 	&	0.218 	&	0.541 	&	64.32 	\\
CogAgent (18B) \cite{hong2024cogagent}	&	0.078 	&	0.278 	&	0.256 	&	64.12 	&	0.243 	&	0.271 	&	0.056 	&	81.25 	&	0.098 	&	0.200 	&	0.224 	&	47.44 	&	0.196 	&	0.344 	&	0.273 	&	50.00 	\\
InternVL3 (38B) \cite{zhu2025internvl3}	&	0.197 	&	0.043 	&	0.642 	&	62.73 	&	0.291 	&	0.227 	&	0.598 	&	82.64 	&	0.320 	&	0.238 	&	0.545 	&	79.33 	&	0.212 	&	0.134 	&	0.644 	&	60.42 	\\
Qwen2.5-VL (72B) \cite{Qwen2.5-VL}	&	0.488 	&	0.227 	&	0.644 	&	75.69 	&	0.576 	&	0.275 	&	0.492 	&	83.10 	&	0.489 	&	0.358 	&	0.520 	&	67.31 	&	0.495 	&	0.267 	&	0.596 	&	70.31 	\\
UI-TARS (72B) \cite{qin2025ui}	&	0.609 	&	0.638 	&	0.526 	&	79.17 	&	0.636 	&	0.599 	&	0.434 	&	81.71 	&	0.642 	&	0.638 	&	0.447 	&	78.37 	&	0.599 	&	0.564 	&	0.531 	&	71.09 	\\
InternVL3 (78B) \cite{zhu2025internvl3}	&	0.416 	&	0.309 	&	0.694 	&	65.05 	&	0.553 	&	0.490 	&	0.547 	&	83.80 	&	0.372 	&	0.403 	&	0.543 	&	81.41 	&	0.448 	&	0.395 	&	0.623 	&	49.91 	\\
\midrule
Qwen2.5-VL (7B) \cite{Qwen2.5-VL}\,\dag	&	\underline{0.865} 	&	\underline{0.820} 	&	\underline{0.798} 	&	84.26 	&	\underline{0.878} 	&	\underline{0.827} 	&	0.702 	&	89.58 	&	\underline{0.854} 	&	0.820 	&	0.698 	&	80.93 	&	\underline{0.853} 	&	\underline{0.809} 	&	\underline{0.750} 	&	79.17 	\\
% UI-TARS (7B) \cite{qin2025ui}\,\dag	&	\underline{0.865} 	&	\underline{0.827} 	&	0.760 	&	84.95 	&	0.872 	&	0.804 	&	0.684 	&	91.20 	&	\underline{0.858} 	&	\underline{0.827} 	&	0.707 	&	\textbf{85.42} 	&	0.857 	&	0.825 	&	0.764 	&	79.25 	\\
InternVL2.5 (8B) \cite{chen2024expanding}\,\dag	&	0.842 	&	0.807 	&	0.768 	&	82.18 	&	0.834 	&	0.765 	&	\underline{0.715} 	&	\textbf{93.98} 	&	0.832 	&	0.773 	&	\underline{0.709} 	&	\underline{83.65} 	&	0.827 	&	0.776 	&	0.749 	&	\textbf{81.86} 	\\
InternVL3 (8B) \cite{zhu2025internvl3}\,\dag	&	0.856 	&	0.817 	&	0.755 	&	\underline{85.19} 	&	0.857 	&	0.777 	&	0.637 	&	\underline{91.90} 	&	0.849 	&	\underline{0.822} 	&	0.680 	&	82.05 	&	0.849 	&	0.799 	&	0.674 	&	\underline{80.64} 	\\
\rowcolor[gray]{.92}
\textbf{MoE-Rater (Ours)}\,*	&	\textbf{0.887} 	&	\textbf{0.859} 	&	\textbf{0.815} 	&	\textbf{85.42} 	&	\textbf{0.895} 	&	\textbf{0.841} 	&	\textbf{0.784} 	&	90.51 	&	\textbf{0.890} 	&	\textbf{0.868} 	&	\textbf{0.745} 	&	\textbf{84.46} 	&	\textbf{0.889} 	&	\textbf{0.865} 	&	\textbf{0.764} 	&	80.38 	\\

\midrule
\midrule
Prompt Category  & \multicolumn{4}{c|}{\textbf{Emotion}}  & \multicolumn{4}{c|}{\textbf{Interaction}}  & \multicolumn{4}{c|}{\textbf{Physical Authenticity}}  & \multicolumn{4}{c}{\textbf{Overall}} \\
\midrule
Method / Metric	&	$\rho_s$\,$\uparrow$	&	$\rho_t$\,$\uparrow$	&	$\rho_c$\,$\uparrow$  &	Acc.\,$\uparrow$	&	$\rho_s$\,$\uparrow$	&	$\rho_t$\,$\uparrow$	&	$\rho_c$\,$\uparrow$  &	Acc.\,$\uparrow$  &	$\rho_s$\,$\uparrow$	&	$\rho_t$\,$\uparrow$	&	$\rho_c$\,$\uparrow$  &	Acc.\,$\uparrow$  &	$\rho_s$\,$\uparrow$	&	$\rho_t$\,$\uparrow$	&	$\rho_c$\,$\uparrow$  &	Acc.\,$\uparrow$ \\
\midrule
Video-ChatGPT (7B) \cite{maaz2023video}	&	-0.014 	&	-0.163 	&	0.041 	&	48.75 	&	-0.047 	&	-0.012 	&	-0.002 	&	47.81 	&	-0.096 	&	-0.050 	&	0.063 	&	69.58 	&	-0.071 	&	-0.080 	&	0.015 	&	60.85 	\\
VideoLLaMA2 (7B) \cite{cheng2024videollama}	&	-0.060 	&	0.033 	&	0.382 	&	48.75 	&	-0.077 	&	-0.074 	&	0.487 	&	47.81 	&	-0.107 	&	-0.069 	&	0.326 	&	69.58 	&	-0.069 	&	-0.048 	&	0.433 	&	60.90 	\\
VideoLLaMA3 (7B) \cite{zhang2025videollama}	&	0.231 	&	0.073 	&	0.416 	&	49.58 	&	0.161 	&	0.064 	&	0.388 	&	52.08 	&	0.213 	&	0.174 	&	0.290 	&	68.19 	&	0.175 	&	0.095 	&	0.363 	&	64.10 	\\
Qwen2-VL (7B) \cite{Qwen2-VL}	&	0.494 	&	0.394 	&	0.522 	&	48.96 	&	0.428 	&	0.415 	&	0.405 	&	48.85 	&	0.451 	&	0.375 	&	0.351 	&	69.03 	&	0.429 	&	0.393 	&	0.405 	&	63.75 	\\
Qwen2.5-VL (7B) \cite{Qwen2.5-VL}	&	0.422 	&	0.384 	&	0.486 	&	71.67 	&	0.412 	&	0.252 	&	0.529 	&	66.98 	&	0.493 	&	0.295 	&	0.467 	&	37.78 	&	0.419 	&	0.307 	&	0.506 	&	63.23 	\\
DeepSeek-VL (7B) \cite{lu2024deepseek}	&	0.102 	&	0.125 	&	0.086 	&	48.33 	&	0.121 	&	0.134 	&	0.003 	&	47.50 	&	0.047 	&	0.089 	&	-0.006 	&	69.44 	&	0.098 	&	0.123 	&	0.008 	&	60.17 	\\
UI-TARS (7B) \cite{qin2025ui}	&	0.415 	&	0.348 	&	0.386 	&	51.67 	&	0.502 	&	0.358 	&	0.447 	&	66.98 	&	0.521 	&	0.301 	&	0.347 	&	69.58 	&	0.495 	&	0.351 	&	0.405 	&	68.83 	\\
InternVL2.5 (8B) \cite{chen2024expanding}	&	0.350 	&	0.291 	&	0.498 	&	48.54 	&	0.452 	&	0.307 	&	0.583 	&	47.81 	&	0.444 	&	0.362 	&	0.398 	&	69.58 	&	0.423 	&	0.323 	&	0.537 	&	60.83 	\\
InternVL3 (8B) \cite{zhu2025internvl3}	&	0.301 	&	0.117 	&	0.456 	&	66.88 	&	0.303 	&	-0.019 	&	0.424 	&	66.67 	&	0.307 	&	-0.004 	&	0.353 	&	43.89 	&	0.268 	&	0.028 	&	0.442 	&	58.31 	\\
InternVL3 (14B) \cite{zhu2025internvl3}	&	0.235 	&	0.285 	&	0.532 	&	70.83 	&	0.202 	&	0.256 	&	0.555 	&	66.98 	&	0.317 	&	0.297 	&	0.409 	&	69.44 	&	0.232 	&	0.261 	&	0.547 	&	68.77 	\\
CogAgent (18B) \cite{hong2024cogagent}	&	0.274 	&	0.324 	&	0.135 	&	48.75 	&	0.183 	&	0.290 	&	0.134 	&	49.17 	&	0.097 	&	0.112 	&	0.042 	&	32.78 	&	0.161 	&	0.254 	&	0.176 	&	50.88 	\\
InternVL3 (38B) \cite{zhu2025internvl3}	&	0.282 	&	0.233 	&	0.578 	&	50.83 	&	0.265 	&	0.144 	&	0.667 	&	49.27 	&	0.334 	&	0.223 	&	0.540 	&	69.44 	&	0.269 	&	0.182 	&	0.622 	&	63.25 	\\
Qwen2.5-VL (72B) \cite{Qwen2.5-VL}	&	0.546 	&	0.302 	&	0.512 	&	72.29 	&	0.503 	&	0.312 	&	0.566 	&	70.63 	&	0.445 	&	0.229 	&	0.419 	&	42.50 	&	0.498 	&	0.284 	&	0.558 	&	67.65 	\\
UI-TARS (72B) \cite{qin2025ui}	&	0.587 	&	0.583 	&	0.466 	&	72.92 	&	0.638 	&	0.596 	&	0.505 	&	66.25 	&	0.626 	&	0.532 	&	0.427 	&	69.17 	&	0.618 	&	0.581 	&	0.501 	&	72.65 	\\
InternVL3 (78B) \cite{zhu2025internvl3}	&	0.470 	&	0.423 	&	0.562 	&	48.75 	&	0.429 	&	0.423 	&	0.679 	&	47.81 	&	0.510 	&	0.468 	&	0.564 	&	69.58 	&	0.450 	&	0.417 	&	0.612 	&	60.83 	\\
\midrule
Qwen2.5-VL (7B) \cite{Qwen2.5-VL}\,\dag	&	\underline{0.850} 	&	\underline{0.814} 	&	\underline{0.768} 	&	76.88 	&	0.870 	&	\underline{0.833} 	&	\underline{0.770} 	&	73.54 	&	\underline{0.854} 	&	0.779 	&	\underline{0.742} 	&	71.81 	&	\underline{0.860} 	&	\underline{0.816} 	&	\underline{0.755} 	&	78.33 	\\
% UI-TARS (7B) \cite{qin2025ui}\,\dag	&	0.831 	&	0.818 	&	0.755 	&	77.71 	&	0.874 	&	0.850 	&	0.758 	&	74.48 	&	0.846 	&	0.784 	&	0.714 	&	74.44 	&	0.859 	&	0.824 	&	0.749 	&	79.81 	\\
InternVL2.5 (8B) \cite{chen2024expanding}\,\dag	&	0.829 	&	0.765 	&	0.763 	&	\textbf{79.38} 	&	0.854 	&	0.816 	&	0.764 	&	\underline{76.88} 	&	0.842 	&	0.775 	&	0.714 	&	\underline{73.47} 	&	0.839 	&	0.785 	&	0.751 	&	\underline{80.71} 	\\
InternVL3 (8B) \cite{zhu2025internvl3}\,\dag	&	0.834 	&	0.810 	&	0.719 	&	77.71 	&	\underline{0.872} 	&	0.828 	&	0.694 	&	74.90 	&	0.849 	&	\underline{0.780} 	&	0.654 	&	72.36 	&	0.855 	&	0.808 	&	0.692 	&	79.56 	\\
\rowcolor[gray]{.92}
\textbf{MoE-Rater (Ours)}\,*	&	\textbf{0.886} 	&	\textbf{0.878} 	&	\textbf{0.799} 	&	\underline{78.54} 	&	\textbf{0.895} 	&	\textbf{0.881} 	&	\textbf{0.810} 	&	\textbf{78.02} 	&	\textbf{0.885} 	&	\textbf{0.849} 	&	\textbf{0.759} 	&	\textbf{79.44} 	&	\textbf{0.891} 	&	\textbf{0.868} 	&	\textbf{0.787} 	&	\textbf{81.48} 	\\

\bottomrule
\end{tabular}
}
\vspace{-2mm}
\label{tab:mllm_compare}
\end{table*}

%% file: tabs/exp_human-agvqa.tex
\begin{table}
\centering
\caption{Comparison with state-of-the-art methods on the Human-AGVQA \cite{zhang2025human} dataset. The best and runner-up performances are bold and underlined, respectively.}
\vspace{-1mm}
\label{tab_benchmark}
\resizebox{0.98\linewidth}{!}{
\begin{tabular}{clcccc}

\toprule
\multirow{2}{*}{\makecell[c]{Quality\\Dimension}}   & \multirow{2}{*}{Method}  & \multicolumn{2}{c}{\textbf{Zero-shot}}  & \multicolumn{2}{c}{\textbf{Fine-tuned}} \\
\cmidrule(r){3-4}   \cmidrule(r){5-6}
&  & SRCC\,$\uparrow$    & PLCC\,$\uparrow$      & SRCC\,$\uparrow$   & PLCC\,$\uparrow$    \\
\midrule                                                                                                                                                                                                              
\multirow{13}{*}{\makecell[c]{Human\\Appearance}} & NIQE \cite{mittal2012making}   & 0.233   & 0.273             & 0.564  & 0.570   \\
                           & BRISQUE \cite{mittal2012no}       & 0.258   & 0.314             & 0.594  & 0.619   \\
                           & CNNIQA \cite{kang2014convolutional}        & 0.243   & 0.296             & 0.620  & 0.646   \\
                           & HyperIQA \cite{su2020blindly}      & 0.318   & 0.339             & 0.684  & 0.696   \\
                           & UNIQUE \cite{zhang2021uncertainty}         & 0.239   & 0.280             & 0.668  & 0.684   \\
                           & MUSIQ \cite{ke2021musiq}         & 0.224   & 0.262             & 0.583  & 0.601   \\
                           & StairIQA \cite{sun2023blind}      & 0.339   & 0.393             & 0.661  & 0.663   \\
                           & CLIP-IQA \cite{wang2023exploring}      & 0.327   & 0.368             & 0.672  & 0.674   \\
                           & LIQE \cite{zhang2023blind}          & 0.283   & 0.332             & 0.640  & 0.656   \\
                           & MA-AGIQA \cite{wang2024large}       & \underline{0.373}   & 0.380             & 0.724  & 0.733 \\
                           & Q-Align \cite{wu2023q}        & 0.362   & \underline{0.419}             & 0.725  & 0.731 \\
                           & GHVQ \cite{zhang2025human}           & -       & -                 & \underline{0.805} & \underline{0.809} \\
                           \rowcolor[gray]{.92}
                           & \textbf{MoE-Rater (Ours)}  & \textbf{0.690}    & \textbf{0.682}    & \textbf{0.860} & \textbf{0.859} \\
\midrule                                                                                                                                                                                                            
\multirow{10}{*}{\makecell[c]{Action\\Continuity}} & ACTION-NET \cite{zeng2020hybrid}   & 0.198   & 0.223            & 0.541  & 0.553   \\
                           & USDL \cite{tang2020uncertainty}                & 0.208   & 0.259             & 0.583  & 0.591   \\
                           & CoRe \cite{yu2021group}                & 0.177   & 0.210             & 0.562  & 0.577   \\
                           & TSA \cite{xu2022finediving}                 & 0.204   & 0.256             & 0.602  & 0.613   \\
                           & Motion Smoothness \cite{huang2024vbench}    & 0.250   & 0.275             & -      & -   \\   
                           & Temporal Flickering \cite{huang2024vbench}  & 0.137   & 0.239             & -      & -   \\
                           & Action-Score \cite{liu2024evalcrafter}         & 0.209   & 0.244             & -      & -   \\
                           & Flow-Score \cite{liu2024evalcrafter}           & \underline{0.254}   & \underline{0.279}             & -      & -   \\
                           & GHVQ \cite{zhang2025human}                 & -       & -                 & \underline{0.771} & \underline{0.778} \\
                           \rowcolor[gray]{.92}
                           & \textbf{MoE-Rater (Ours)}  & \textbf{0.656}    & \textbf{0.650}    & \textbf{0.830} & \textbf{0.838} \\
\midrule                                                                                                                                                                                                                  
\multirow{14}{*}{\makecell[c]{Overall\\Video}} & TLVQM \cite{korhonen2019two}          & 0.272   & 0.312             & 0.603  & 0.617   \\
                           & RAPIQUE \cite{tu2021rapique}              & 0.313   & 0.351             & 0.621  & 0.637   \\
                           & VIDEAL \cite{tu2021ugc}              & 0.342   & 0.353             & 0.628  & 0.642   \\
                           & PatchVQ \cite{ying2021patch}             & 0.379   & 0.399             & 0.657  & 0.693   \\
                           & SimpleVQA \cite{sun2022deep}            & 0.364   & 0.378             & 0.687  & 0.698   \\
                           & BVQA \cite{li2022blindly}                & 0.349   & 0.374             & 0.672  & 0.702   \\
                           & FastVQA \cite{wu2022fast}              & 0.390   & 0.412             & 0.698  & 0.711   \\
                           & DOVER \cite{wu2023exploring}               & 0.312   & 0.337             & 0.705  & 0.717   \\
                           & T2VQA \cite{kou2024subjective}                & 0.359   & 0.367             & 0.737  & 0.742 \\
                           & UGVQ \cite{zhang2024benchmarking}                & 0.349   & 0.358             & 0.734  & 0.743 \\
                           & EvalCrafter \cite{liu2024evalcrafter}          & 0.328   & 0.336             & -      & -        \\
                           & Q-Align \cite{wu2023q}              & \underline{0.422}   & \underline{0.481}             & 0.715  & 0.723        \\
                           & GHVQ \cite{zhang2025human}                & -       & -                 & \underline{0.768}  & \underline{0.773} \\
                           \rowcolor[gray]{.92}
                           & \textbf{MoE-Rater (Ours)}  & \textbf{0.757}    & \textbf{0.751}    & \textbf{0.823} & \textbf{0.824} \\
\bottomrule
\end{tabular}
}
\vspace{-3mm}
\label{tab:exp_human-agvqa}
\end{table}

%% file: tabs/t2v_model_rank.tex
\begin{table}[t]
\centering
% \belowrulesep=0pt
% \aboverulesep=0pt
% \renewcommand\arraystretch{1}
\caption{Evaluation of 24 T2V generation models across four evaluation dimensions, with an overall score obtained by averaging these dimensions. The abbreviations ``Corr.", ``H.", and ``M." denote text-video correspondence, human-evaluation scores, and scores predicted by our MoE-Rater model, respectively. All scores are reported on a 0-100 scale, with higher values indicating better performance. The t2v models are ranked by their overall scores from human evaluation.}
\vspace{-1mm}

% \resizebox{\linewidth}{!}{
% \begin{tabular}{l|cc|cc|cc|cc|cc}
% \toprule
% Dimension  & \multicolumn{2}{c|}{\textbf{Spatial}}  & \multicolumn{2}{c|}{\textbf{Temporal}}  & \multicolumn{2}{c|}{\textbf{Corr.}}  & \multicolumn{2}{c|}{\textbf{Q\&A}}  & \multicolumn{2}{c}{\textbf{Overall}} \\
% \hdashline

\resizebox{\linewidth}{!}{
\begin{tabular}{lc>{\columncolor[gray]{.92}}cc>{\columncolor[gray]{.92}}cc>{\columncolor[gray]{.92}}cc>{\columncolor[gray]{.92}}cc>{\columncolor[gray]{.92}}c}
\toprule
Dimension  & \multicolumn{2}{c}{Spatial}  & \multicolumn{2}{c}{Temporal}  & \multicolumn{2}{c}{Corr.}  & \multicolumn{2}{c}{Q\&A}  & \multicolumn{2}{c}{Overall} \\ \cmidrule(lr){2-3} \cmidrule(lr){4-5} \cmidrule(lr){6-7} \cmidrule(lr){8-9} \cmidrule(lr){10-11}

Model	&	H.	&	M.	&	H.	&	M.	&	H.	&	M.	&	H.	&	M.	&	H.	&	M.	\\
\midrule

Wan2.1-14B \cite{wan2025wan}	&	64.9 	&	65.4 	&	66.6 	&	62.7 	&	60.7 	&	59.3 	&	68.5 	&	59.0 	&	65.2 	&	61.6 	\\
PixVerse-V4	\cite{PixVerse}	&	67.1 	&	67.7 	&	68.0 	&	66.4 	&	62.0 	&	61.3 	&	63.5 	&	63.0 	&	65.1 	&	64.6 	\\
SkyReels-V2 \cite{chen2025skyreels}	&	61.9 	&	62.8 	&	63.5 	&	61.5 	&	58.2 	&	57.4 	&	58.5 	&	58.5 	&	60.5 	&	60.1 	\\
Dreamina \cite{Dreamina}	&	67.1 	&	70.2 	&	67.7 	&	67.7 	&	57.2 	&	57.0 	&	50.0 	&	52.0 	&	60.5 	&	61.7 	\\
Sora \cite{sora}	&	64.5 	&	66.9 	&	64.0 	&	65.0 	&	56.9 	&	58.3 	&	50.0 	&	56.0 	&	58.9 	&	61.6 	\\
HunyuanVideo \cite{kong2024hunyuanvideo}	&	62.5 	&	63.8 	&	62.5 	&	62.9 	&	56.9 	&	57.5 	&	52.5 	&	55.0 	&	58.6 	&	59.8 	\\
Pika-2.2 \cite{pika}	&	56.6 	&	59.2 	&	54.0 	&	57.4 	&	59.6 	&	60.1 	&	62.0 	&	62.5 	&	58.0 	&	59.8 	\\
Kling-1.0 \cite{kling}	&	59.2 	&	61.2 	&	63.2 	&	63.2 	&	55.6 	&	55.8 	&	54.0 	&	56.0 	&	58.0 	&	59.0 	\\
Wan2.1-1.3B \cite{wan2025wan}	&	61.6 	&	63.0 	&	59.1 	&	60.1 	&	56.2 	&	56.5 	&	54.5 	&	51.5 	&	57.8 	&	57.8 	\\
Gen-3 \cite{gen3}	&	62.2 	&	65.5 	&	61.0 	&	63.8 	&	50.6 	&	52.7 	&	37.5 	&	45.0 	&	52.9 	&	56.7 	\\
Step-Video-T2V \cite{ma2025step}	&	55.8 	&	58.6 	&	59.3 	&	59.8 	&	51.6 	&	53.0 	&	44.5 	&	43.5 	&	52.8 	&	53.7 	\\
MAGI-1 \cite{teng2025magi}	&	53.3 	&	55.1 	&	54.7 	&	55.6 	&	51.7 	&	52.8 	&	40.5 	&	42.5 	&	50.1 	&	51.5 	\\
CogVideoX1.5 \cite{yang2024cogvideox}	&	49.0 	&	51.3 	&	46.1 	&	49.0 	&	52.7 	&	53.4 	&	44.0 	&	45.0 	&	48.0 	&	49.7 	\\
Mochi \cite{genmo2024mochi}	&	43.4 	&	45.1 	&	47.1 	&	47.8 	&	52.3 	&	53.1 	&	46.5 	&	45.5 	&	47.3 	&	47.9 	\\
VideoCrafter2 \cite{chen2024videocrafter2}	&	43.0 	&	43.9 	&	44.2 	&	42.4 	&	46.3 	&	45.3 	&	29.0 	&	21.5 	&	40.6 	&	38.3 	\\
AnimateLCM \cite{wang2024animatelcm}	&	49.3 	&	52.1 	&	48.3 	&	49.0 	&	43.6 	&	43.6 	&	21.0 	&	16.5 	&	40.6 	&	40.3 	\\
AnimateDiff \cite{lin2024animatediff}	&	46.4 	&	48.5 	&	45.4 	&	45.0 	&	41.9 	&	42.2 	&	21.0 	&	14.5 	&	38.7 	&	37.5 	\\
T2V-Turbo-V2 \cite{li2024t2v}	&	42.9 	&	45.2 	&	40.7 	&	40.7 	&	43.6 	&	44.3 	&	25.0 	&	22.0 	&	38.0 	&	38.0 	\\
Latte \cite{ma2024latte}	&	40.8 	&	42.8 	&	38.8 	&	40.4 	&	41.8 	&	42.7 	&	21.5 	&	16.5 	&	35.7 	&	35.6 	\\
Show-1 \cite{zhang2024show}	&	36.1 	&	36.3 	&	39.1 	&	37.8 	&	43.9 	&	43.9 	&	23.5 	&	25.0 	&	35.6 	&	35.7 	\\
MagicTime \cite{yuan2025magictime}	&	40.8 	&	42.2 	&	39.4 	&	40.0 	&	38.9 	&	39.2 	&	16.0 	&	10.5 	&	33.8 	&	33.0 	\\
ZeroScope \cite{Zeroscope}	&	34.3 	&	34.4 	&	36.7 	&	36.2 	&	42.1 	&	41.9 	&	21.0 	&	16.5 	&	33.5 	&	32.3 	\\
ModelScope \cite{wang2023modelscope}	&	28.6 	&	28.0 	&	33.8 	&	32.4 	&	39.3 	&	39.0 	&	20.0 	&	14.5 	&	30.4 	&	28.5 	\\
LTX-Video \cite{hacohen2024ltx}	&	32.5 	&	33.6 	&	36.1 	&	36.2 	&	35.8 	&	37.1 	&	14.5 	&	13.0 	&	29.7 	&	30.0 	\\
% \hdashline
\midrule

SRCC to Human	&	-	&	\textbf{0.99} 	&	-	&	\textbf{0.97} 	&	-	&	\textbf{0.98} 	&	-	&	\textbf{0.97} 	&	-	&	\textbf{0.99} 	\\
PLCC to Human	&	-	&	\textbf{1.00} 	&	-	&	\textbf{0.99} 	&	-	&	\textbf{0.99} 	&	-	&	\textbf{0.98} 	&	-	&	\textbf{0.99} 	\\

\bottomrule
\end{tabular}
}
\vspace{-3mm}
\label{tab:t2v_compare}
\end{table}

%% file: tabs/ablation.tex
\begin{table*}[t]
\centering
% \belowrulesep=0pt
% \aboverulesep=0pt
% \renewcommand\arraystretch{1}
\caption{Ablation study of MoE-Rater on the HVEval+ dataset. The abbreviation ``Acc." below ``Q\&A" represents category-specific question-answering accuracy, while in all other columns it denotes pairwise comparison accuracy. The best and runner-up performances are bold and underlined, respectively.}
\vspace{-2mm}

\resizebox{\linewidth}{!}{
\begin{tabular}{cccccccc||ccc|ccc|ccc|c}
\toprule
\multicolumn{8}{c||}{\textbf{Feature \& Strategy}}  & \multicolumn{3}{c|}{\textbf{Spatial Quality}}  & \multicolumn{3}{c|}{\textbf{Temporal Quality}}  & \multicolumn{3}{c|}{\textbf{Text-Video Correspondence}}  & \textbf{Q\&A}  \\

% \noalign{\vskip 1pt}
% \hdashline
% \noalign{\vskip 1pt}
\midrule
% \hline

% Spatial	&	Temporal	&	ViT LoRA	&	LLM LoRA	&	Task-Specific Adaptation	&	MoPE	&	MoLE	&	SRCC	&	PLCC	&	Acc	&	SRCC	&	PLCC	&	Acc	&	SRCC	&	PLCC	&	Acc	&	Acc	\\
\multirow{2}{*}{No.}  &	\multirow{2}{*}{\makecell[c]{Spatial\\[-1pt]Feature}}	&	\multirow{2}{*}{\makecell[c]{Temporal\\[-1pt]Feature}}	&	\multirow{2}{*}{\makecell[c]{ViT\\[-1pt]LoRA}}	&	\multirow{2}{*}{\makecell[c]{LLM\\[-1pt]LoRA}}	&	\multirow{2}{*}{\makecell[c]{Task-Specific\\[-1pt]Adaptation}}	&	\multirow{2}{*}{MoPE}	&	\multirow{2}{*}{MoLE}	&	\multirow{2}{*}{SRCC$\uparrow$}	&	\multirow{2}{*}{PLCC$\uparrow$}	&	\multirow{2}{*}{Acc.$\uparrow$}	&	\multirow{2}{*}{SRCC$\uparrow$}	&	\multirow{2}{*}{PLCC$\uparrow$}	&	\multirow{2}{*}{Acc.$\uparrow$}	&	\multirow{2}{*}{SRCC$\uparrow$}	&	\multirow{2}{*}{PLCC$\uparrow$}	&	\multirow{2}{*}{Acc.$\uparrow$}	&	\multirow{2}{*}{\centering Acc.$\uparrow$}	\\
& & & & & & & & & & & & & & & & \\

% \hline
% \noalign{\vskip 1pt}
\midrule

(1)	&	\checkmark	&		&		&		&		&		&		&	0.8634 	&	0.8655 	&	95.15\%	&	0.8382 	&	0.8341 	&	93.50\%	&	0.7349 	&	0.7259 	&	87.85\%	&	78.35\%	\\
(2)	&	\checkmark	&	\checkmark	&		&		&		&		&		&	0.8687 	&	0.8571 	&	95.33\%	&	0.8426 	&	0.8406 	&	94.04\%	&	0.7406 	&	0.7260 	&	88.63\%	&	78.85\%	\\
(3)	&	\checkmark	&	\checkmark	&	\checkmark	&		&		&		&		&	0.8864 	&	0.8907 	&	96.63\%	&	0.8543 	&	0.8549 	&	95.06\%	&	0.7526 	&	0.7425 	&	89.69\%	&	79.79\%	\\
(4)	&	\checkmark	&	\checkmark	&		&	\checkmark	&		&		&		&	0.8769 	&	0.8726 	&	95.56\%	&	0.8428 	&	0.8420 	&	94.31\%	&	0.7610 	&	0.7542 	&	89.17\%	&	79.96\%	\\
(5)	&	\checkmark	&	\checkmark	&	\checkmark	&	\checkmark	&		&		&		&	0.8875 	&	0.8908 	&	\underline{96.67\%}	&	0.8615 	&	0.8639 	&	95.15\%	&	0.7649 	&	0.7464 	&	89.75\%	&	79.38\%	\\
% (6)	&	\checkmark	&	\checkmark	&	\checkmark	&	\checkmark	&		&	\checkmark	&		&	0.8878 	&	0.8915 	&	\textbf{96.71\%}	&	0.8610 	&	0.8639 	&	95.19\%	&	0.7650 	&	0.7467 	&	89.79\%	&	79.31\%	\\
% (7)	&	\checkmark	&	\checkmark	&	\checkmark	&	\checkmark	&		&		&	\checkmark	&	0.8878 	&	0.8915 	&	\textbf{96.71\%}	&	0.8613 	&	0.8633 	&	95.19\%	&	0.7644 	&	0.7461 	&	89.85\%	&	79.31\%	\\
(6)	&	\checkmark	&	\checkmark	&	\checkmark	&	\checkmark	&		&	\checkmark	&	\checkmark	&	0.8878 	&	0.8915 	&	\textbf{96.71\%}	&	0.8607 	&	0.8635 	&	95.19\%	&	0.7644 	&	0.7461 	&	89.85\%	&	79.31\%	\\
(7)	&	\checkmark	&	\checkmark	&	\checkmark	&	\checkmark	&	\checkmark	&		&		&	0.8894 	&	0.8957 	&	96.33\%	&	0.8660 	&	0.8683 	&	\underline{95.40\%}	&	\underline{0.7823} 	&	0.7748 	&	89.63\%	&	81.17\%	\\
(8)	&	\checkmark	&	\checkmark	&	\checkmark	&	\checkmark	&	\checkmark	&	\checkmark	&		&	0.8910 	&	0.8963 	&	\underline{96.67\%}	&	\underline{0.8671} 	&	\underline{0.8694} 	&	95.25\%	&	0.7817 	&	\underline{0.7760} 	&	\underline{90.50\%}	&	\textbf{81.60\%}	\\
(9)	&	\checkmark	&	\checkmark	&	\checkmark	&	\checkmark	&	\checkmark	&		&	\checkmark	&	\textbf{0.8927} 	&	\textbf{0.8986} 	&	96.63\%	&	0.8654 	&	0.8680 	&	95.25\%	&	0.7822 	&	0.7751 	&	90.46\%	&	81.15\%	\\

\rowcolor[gray]{.92}
(10)	&	\checkmark	&	\checkmark	&	\checkmark	&	\checkmark	&	\checkmark	&	\checkmark	&	\checkmark	&	\underline{0.8911} 	&	\underline{0.8972} 	&	96.56\%	&	\textbf{0.8676} 	&	\textbf{0.8704} 	&	\textbf{95.63\%}	&	\textbf{0.7871} 	&	\textbf{0.7795} 	&	\textbf{90.54\%}	&	\underline{81.48\%}	\\

\bottomrule
\end{tabular}
}
\vspace{-3mm}
\label{tab:ablation}
\end{table*}